\documentclass[10pt,twocolumn,letterpaper]{article}

\usepackage[pagenumbers]{cvpr}
\usepackage{amsmath}

\definecolor{cvprblue}{rgb}{0.21,0.49,0.74}
\usepackage[pagebackref,breaklinks,colorlinks,allcolors=cvprblue]{hyperref}

\usepackage{amsmath,amsfonts,bm}

\def\eqref#1{equation~\ref{#1}}

\def\1{\bm{1}}

\def\va{{\bm{a}}}

\def\vq{{\bm{q}}}

\def\mA{{\bm{A}}}

\def\mK{{\bm{K}}}

\def\mQ{{\bm{Q}}}

\def\mV{{\bm{V}}}

\def\mX{{\bm{X}}}

\def\mZ{{\bm{Z}}}

\DeclareMathAlphabet{\mathsfit}{\encodingdefault}{\sfdefault}{m}{sl}
\SetMathAlphabet{\mathsfit}{bold}{\encodingdefault}{\sfdefault}{bx}{n}

\usepackage{amsmath,amsfonts,amssymb,amsthm,mathtools}
\usepackage{multirow}
\usepackage{tabularx}
\usepackage{xcolor}
\usepackage[dvipsnames]{xcolor}
\usepackage{colortbl}
\usepackage{kotex}
\usepackage{pifont} 
\usepackage{makecell}

\newcommand{\appdx}[1]{\textcolor{black}{#1}}
\newcommand{\cmark}{\textcolor{black}{\ding{51}}}

\title{Your Large Vision-Language Model Only Needs 

A Few Attention Heads For Visual Grounding}

\author{
Seil Kang\quad Jinyeong Kim\quad Junhyeok Kim\quad Seong Jae Hwang \\
{Yonsei University}\\
$\mathtt{\small \{seil,jinyeong1324,timespt,seongjae\}@yonsei.ac.kr}$\\
}

\begin{document}
\maketitle
\label{sec:abs}
\begin{abstract}
Visual grounding seeks to localize the image region corresponding to a free-form text description. Recently, the strong multimodal capabilities of Large Vision-Language Models (LVLMs) have driven substantial improvements in visual grounding, though they inevitably require fine-tuning and additional model components to explicitly generate bounding boxes or segmentation masks. However, we discover that a few attention heads in frozen LVLMs demonstrate strong visual grounding capabilities. We refer to these heads, which consistently capture object locations related to text semantics, as localization heads. Using localization heads, we introduce a straightforward and effective training-free visual grounding framework that utilizes text-to-image attention maps from localization heads to identify the target objects. Surprisingly, only three out of thousands of attention heads are sufficient to achieve competitive localization performance compared to existing LVLM-based visual grounding methods that require fine-tuning. Our findings suggest that LVLMs can innately ground objects based on a deep comprehension of the text-image relationship, as they implicitly focus on relevant image regions to generate informative text outputs. All the source codes will be made available to the public.
    
\vspace{-5pt}
\end{abstract}
\section{Introduction}
\label{sec:1}
Visual grounding is a task that, given textual descriptions, identifies and localizes relevant objects within an image, producing outputs such as bounding boxes~\cite{mao2016rec1, yu2016rec2} or segmentation masks~\cite{hu2016res1}.
Recently, this vision-language task, which inherently requires a deep understanding of the relationship between images and text, has seen significant advancements with the emergence of powerful Large Vision-Language Models (LVLMs)~\cite{LLaVA, LLaVA1.5, BLIP2, Cambrian1}.
However, since LVLMs are primarily designed to generate text outputs, directly leveraging them as a vision-language tool to identify and localize objects within an image (\ie, visual grounding) presents technical challenges.
Inevitably, current LVLM-based visual grounding methods require explicit fine-tuning of LVLMs with additional visual grounding datasets and modifications to model components to enable the generation of bounding boxes~\cite{chen2023shikra, you2023ferret, wang2023cogvlm} or segmentation masks~\cite{lai2024lisa, rasheed2024glamm, zhang2024psalm, xia2024gsva}.

\begin{figure}[t]
    \centering
    \includegraphics[width=\linewidth]{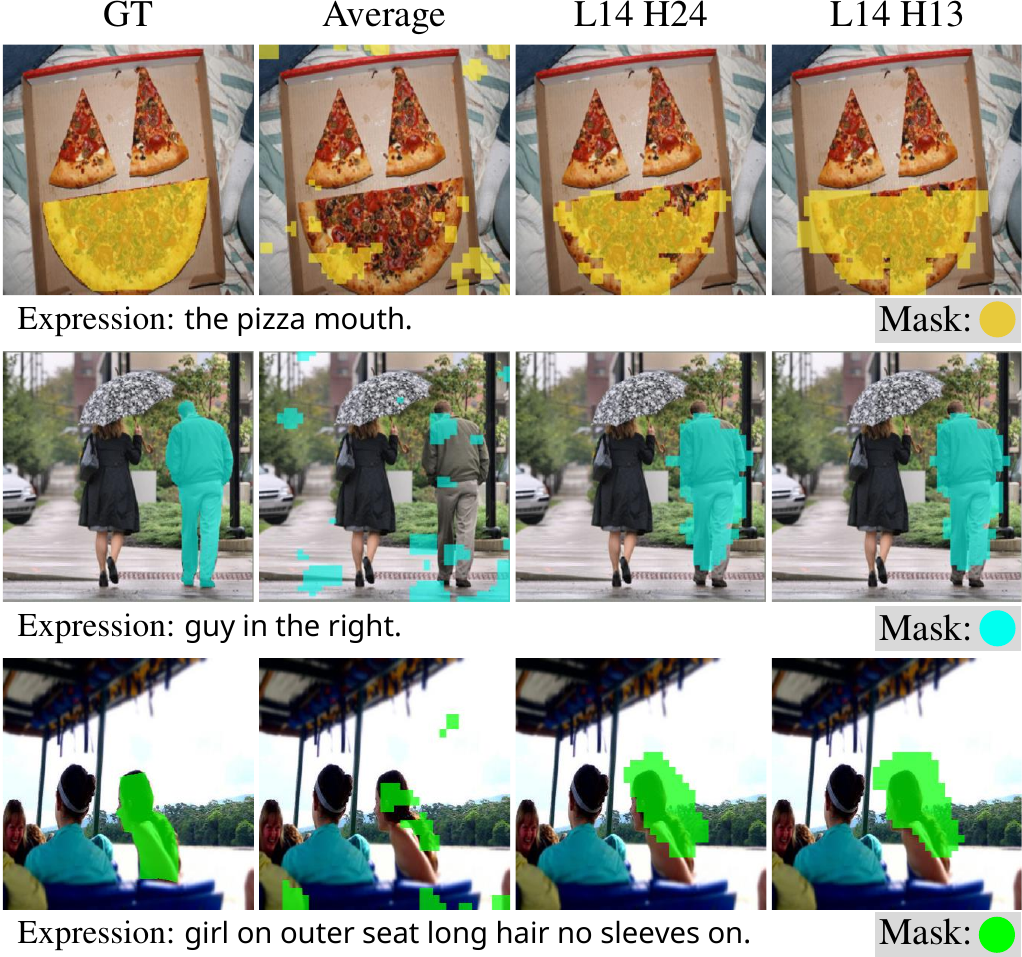}
    \vspace{-15pt}
    \caption{Visualization of the text-to-image attention maps from LLaVA-1.5-7B~\cite{LLaVA1.5}. While the average attention map initially seems uninformative for localization, a closer examination reveals that LVLM possesses built-in \textit{localization heads} that consistently capture key areas of an image corresponding to the referring text, regardless of sample variations. L14 H24 refers to the 24th attention head in the 14th layer of the LVLM.}
    \vspace{-15pt}
    \label{fig:1}
\end{figure}

Despite the interesting integration of LVLMs in previous visual grounding works, a fundamental question remains: \textit{since LVLMs generate text outputs that imply an understanding of specific image regions, is it possible to explicitly observe this mechanism in action?}
In other words, we ask whether we can extract how the LVLMs ``focus" on specific image regions corresponding to given text descriptions for visual grounding.
A natural first approach to addressing this question might be to examine the text-to-image attention maps, which reveal how a text description attends to different image patches. To explore this, we visualize the average attention maps of LVLMs across various layers and heads---a common method in ViTs~\cite{zhang2022segvit, gao2021ts} and diffusion models (DMs)~\cite{tang2022daam, hertz2022prompt, cao2023masactrl}---anticipating that they would capture the regions associated with the referring text. However, unlike the interpretable attention patterns observed in ViTs and DMs, the text-to-image attention maps in LVLMs appear sparse and contain significant noise, as illustrated in the second column of Fig.~\ref{fig:1}. This suggests that the current use of LVLM attention maps may struggle to accurately pinpoint relevant objects for visual grounding.

However, interestingly, our work reveals that not the average of the attention maps, but \textit{some} small subset of attention heads are capable of providing tangible and precise text-image attention maps.
In particular, we find that a few attention heads in LVLMs consistently capture regions in images corresponding to the referred text, regardless of the samples.
We refer to these heads as \textit{localization heads}.
For example, as presented in the third and fourth columns of Fig.~\ref{fig:1}, the attention maps of the 24th head of the 14th layer (L14 H24) and the 13th head of the 14th layer (L14 H13) in LLaVA-1.5-7B~\cite{LLaVA1.5} consistently highlight the regions of interest based on the referred text.

In this work, we introduce how we systematically identify such localization heads based on two explicit criteria.
(1) We measure how much each attention head focuses on the image by calculating the \textit{attention sum} and only select the heads that dominantly attend to the image. 
(2) Among these heads, the ones that specifically pay attention to a certain region of the image, which is measured by \textit{spatial entropy}~\cite{batty1974spatial}, are considered to effectively localize the referred object.
We validate that the selected localization heads consistently capture objects closely associated with the text.

\begin{figure}[t]
    \centering
    \includegraphics[width=\linewidth]{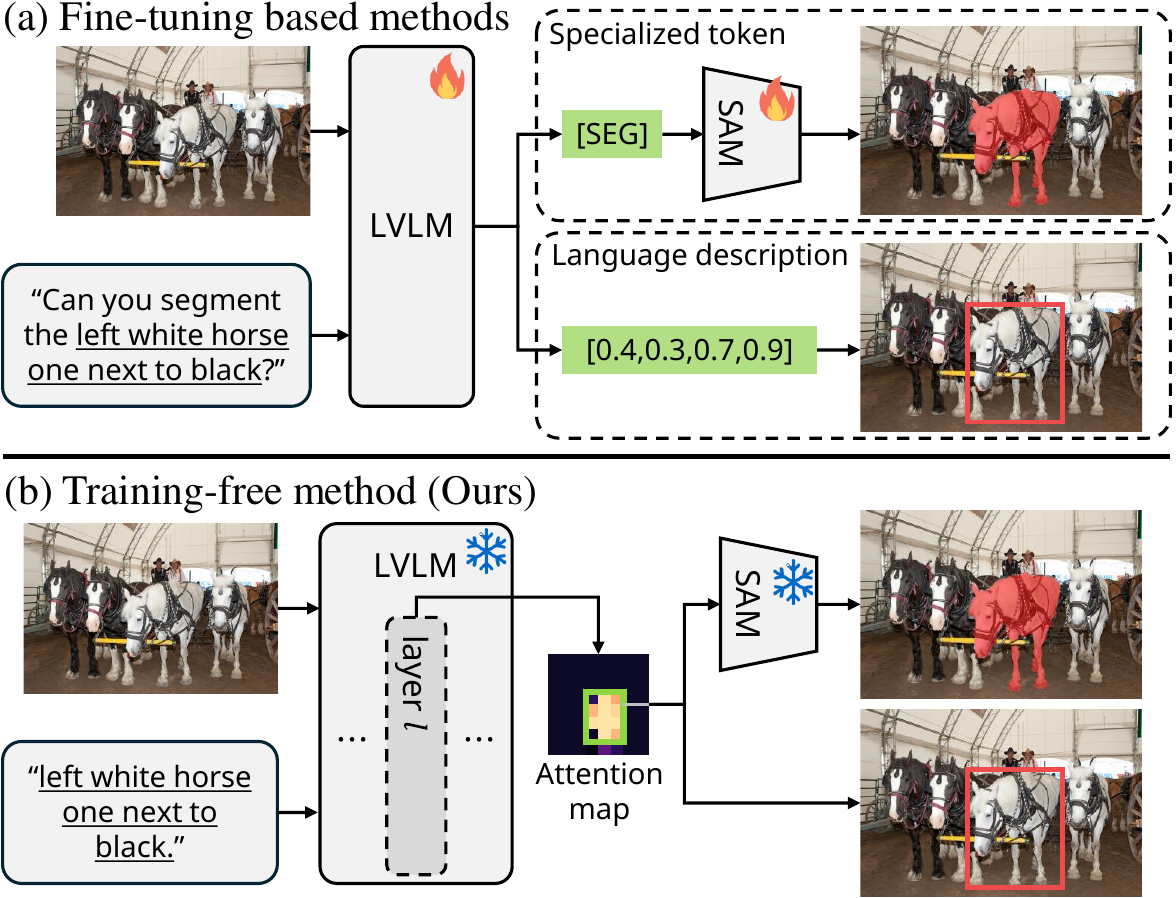}
    \vspace{-15pt}
    \caption{Comparison of LVLM frameworks for visual grounding. (a) Existing methods generally fine-tune a LVLM to leverage specialized tokens (\eg, $\mathtt{[SEG]}$) or language descriptions for visual grounding. (b) Our framework utilizes the attention maps of only a few localization heads from frozen LVLMs.}
    \vspace{-10pt}
    \label{fig:2}
\end{figure}

With our localization heads, we introduce a simple yet effective training-free visual grounding framework.
The attention maps from the localization heads are assembled to predict the bounding box or mask of the referred object.
Notably, only three localization heads are enough to localize the referred object within the image, suggesting that they are highly specialized to attend to relevant image regions.
As shown in Fig.~\ref{fig:2}, in contrast to existing fine-tuning based methods, our framework is training-free, eliminating the need for additional fine-tuning LVLMs for visual grounding tasks.

We validate our approach across ten different LVLMs with varying parameter counts, architectures, and training datasets, demonstrating its broad applicability.
Our framework outperforms the existing training-free methods by significant margins. 
Furthermore, our method performs comparably to specially fine-tuned LVLMs for visual grounding tasks (\eg, LISA~\cite{lai2024lisa}).
The results indicate that LVLMs can serve as effective text-referring localizers, intrinsically identifying regions that are relevant and coherent with the text expression.
To the best of our knowledge, we are the first to identify the localization properties of specific attention heads in LVLMs. 

In summary, our contributions are as follows:
\begin{itemize}
    \item We discover that the specific attention heads in LVLMs have the capability for visual grounding, which we refer to as \textit{localization heads}.
    \item We propose a simple yet effective framework for LVLM-based training-free visual grounding with localization heads. The attention maps from a few localization heads are utilized to predict the bounding box or mask of the referred object.
    \item We evaluate our approach across various LVLMs. Our framework demonstrates superior performance by a large margin compared to other training-free methods and even performs comparably to fine-tuned methods.
\end{itemize}

\section{Related Works}
\label{sec:2}
\noindent\textbf{Visual Grounding.} Visual grounding aims to identify the region in the image based on a free-form natural language expression~\cite{chandu2021grounding}, which has expanded the scope of detection and segmentation tasks to a more realistic scenario~\cite{shridhar2020ingress, xu2023meta}. Two prominent tasks within visual grounding are Referring Expression Comprehension (REC)~\cite{mao2016rec1, yu2016rec2} and Referring Expression Segmentation (RES)~\cite{hu2016res1, liu2023gres}. REC focuses on localizing a referred object in an image and generating a bounding box, while RES further requires a pixel-level segmentation mask. In order to address these tasks, numerous studies have been conducted to explore effective methods that consider both text and visual information simultaneously~\cite{liu2017classic1, li2018classic2, shi2018classic3, yang2022lavt, wang2022cris, kim2024risclip, zhao2023VPD, pnvr2023LDZNet}.

\noindent\textbf{Application of LVLMs in Grounding Tasks.} Recently, visual grounding has been significantly advanced by leveraging the outstanding vision-language processing capabilities of LVLMs. To incorporate LVLMs into visual grounding tasks, existing methods include visual grounding datasets in the training process and implement additional components to extract localization information. For example, LISA~\cite{lai2024lisa} introduces \texttt{[SEG]} token as a mask embedding and generates a segmentation mask using additional mask decoder~\cite{SAM}. F-LMM~\cite{wu2024flmm} leverages the attention weights of frozen LVLMs, but still requires training its mask refinement modules on visual grounding datasets. In contrast, we propose a training-free visual grounding method that directly utilizes LVLMs.

\vspace{3pt}

\noindent\textbf{Training-Free Visual Grounding.} 
Given the high performance of multimodal foundation models across diverse vision-language tasks, training-free visual grounding emerges as a new research direction.
Existing training-free methods typically apply internal features or attention maps from CLIP~\cite{CLIP} or Text-to-Image Diffusion Models (DMs)~\cite{StableDiffusion}.
CLIP-based methods typically employ off-the-shelf models~\cite{FasterRCNN, SAM} to generate region proposals and select the most relevant bounding box~\cite{subramanian2022ReCLIP} or mask~\cite{yu2023ZSRS, suo2023TAS} based on the CLIP similarity score with the text query.
On the other hand, DM-based methods utilize the residue of the text-to-image diffusion process (\eg, the attention map) to predict the segmentation mask~\cite{burgert2022Peekaboo, ni2023RefDiff}.
Our work advances this line of research by introducing the first LVLM-based training-free visual grounding framework.
\section{Background}
\label{sec:3}
\noindent\textbf{Notation.} Large Vision-Language Models (LVLMs) typically consist of three main components: a vision encoder, a projector, and a large language model. For an input image $\mX_\text{v}$, the vision encoder and the projector transform the image into a sequence of visual embedding $\mZ_\text{v} \in \mathbb{R}^{P^2 \times d}$, where $P^2$ is the number of flattened image tokens and $d$ is the hidden dimension. Similarly, an input text $\mX_\text{t}$ is converted into a sequence of token embeddings $\mZ_\text{t} \in \mathbb{R}^{L \times d}$, where $L$ is the number of tokens in the text. The visual and textual embeddings are concatenated as $\mZ^0 = [\mZ_\text{v}; \mZ_\text{t}] \in \mathbb{R}^{(P^2 + L) \times d}$ and fed into the large language model (LLM) as the input embeddings.

\vspace{3pt}
\noindent\textbf{Multi-Head Self-Attention.} The input embeddings $\mZ^0$ pass through a series of decoder blocks, which consists of multi-head self-attention and feed-forward neural network module. Specifically, we focus on the attention heads, as these are the only components where tokens interact. In layer $\ell$ and head $h$, the hidden state from the previous layer $\mZ^{\ell-1}$ is projected into query $\mQ$, key $\mK$, and value $\mV \in \mathbb{R}^{(P^2 + L) \times d_h}$ matrices, where $d_h$ is the hidden dimension of the attention head. Then, the attention head computes the attention weights as:
\begin{equation}
\label{eq:prelim-attn}
    \text{Attn}^{\ell,h}(\mZ^{\ell-1}) = \text{softmax}\left(\frac{\mQ \mK^\top}{\sqrt{d_h}}\right).
\end{equation}
Note that the attention weights reflect the similarity between the query $\mQ$ and key $\mK$ matrices.

\vspace{3pt}
\noindent\textbf{Investigation of Image-Text Interaction.} Considering that LLM decoding operates in an auto-regressive manner, information flows from preceding tokens to subsequent ones, resulting in the final token to encapsulate the context of the entire sentence~\cite{wang2023label, jiang2024interpreting}. Thus, we posit that the query vector of the last input text token $\vq_\text{txt}$ serves as a representative query for the whole sentence. For example, in the sentence ``$\mathtt{the\ pizza\ mouth.}$'' in \cref{fig:1}, the query vector of the last token [$\mathtt{.}$] is utilized in our experiments. To investigate image-text interactions, we examine the attention weights of where the query is $\vq_\text{txt}$ and keys are image tokens. Specifically, considering a slight modification of Eq.~(\ref{eq:prelim-attn}), for the attention weights $\va^{\ell,h}$ at layer $\ell$ and head $h$ with $\vq_\text{txt}$ as a query token:
\begin{equation}
    \va^{\ell,h} = \text{softmax}\left(\frac{\vq_\text{txt} \mK^\top}{\sqrt{d_h}}\right) \in \mathbb{R}^{P^2 + L},
\end{equation}
we focus on the first $P^2$ components, $\va^{\ell,h}[1:P^2]$, for our analysis. In the following sections of this paper, this will also be denoted as L$\ell$ H$h$ for simplicity. For example, L5 H3 refers to the third attention head in the fifth layer of the LVLM.

\section{Towards Discovering Localization Heads}
\label{sec:4}
Recent studies~\cite{clark2019bert, voita2019analyzing, zheng2024attention} have shown that the attention heads exhibit distinct characteristics, motivating us to find specific heads possessing the potential to serve as effective \textit{text referring localizers}. In this section, we propose attention sum and spatial entropy in \cref{sec:4_1} as two criteria for selecting such heads. Through experiments in \cref{sec:4_2}, we validate that the heads capturing objects corresponding to the text description can be successfully identified based on the proposed criteria. Note that the first two layers of the LLM are consistently excluded in our analyses, as the early layers are known to operate differently from the other layers~\cite{stageofinference}. To demonstrate the generalizability of our findings, we conduct experiments across various LVLMs~\cite{lu2024deepseek, li2024mini-gemini, chen2024internvl, young2024yi, chen2023sharegpt4v, LLaVA, LLaVA1.5} and datasets~\cite{kazemzadeh2014referitgame, hu2016res1}. Details of the experimental setup and more results are provided in the \appdx{Appendix~Sec.~A. and C.}, respectively.

\begin{figure}[t]
    \centering
    \includegraphics[width=\linewidth]{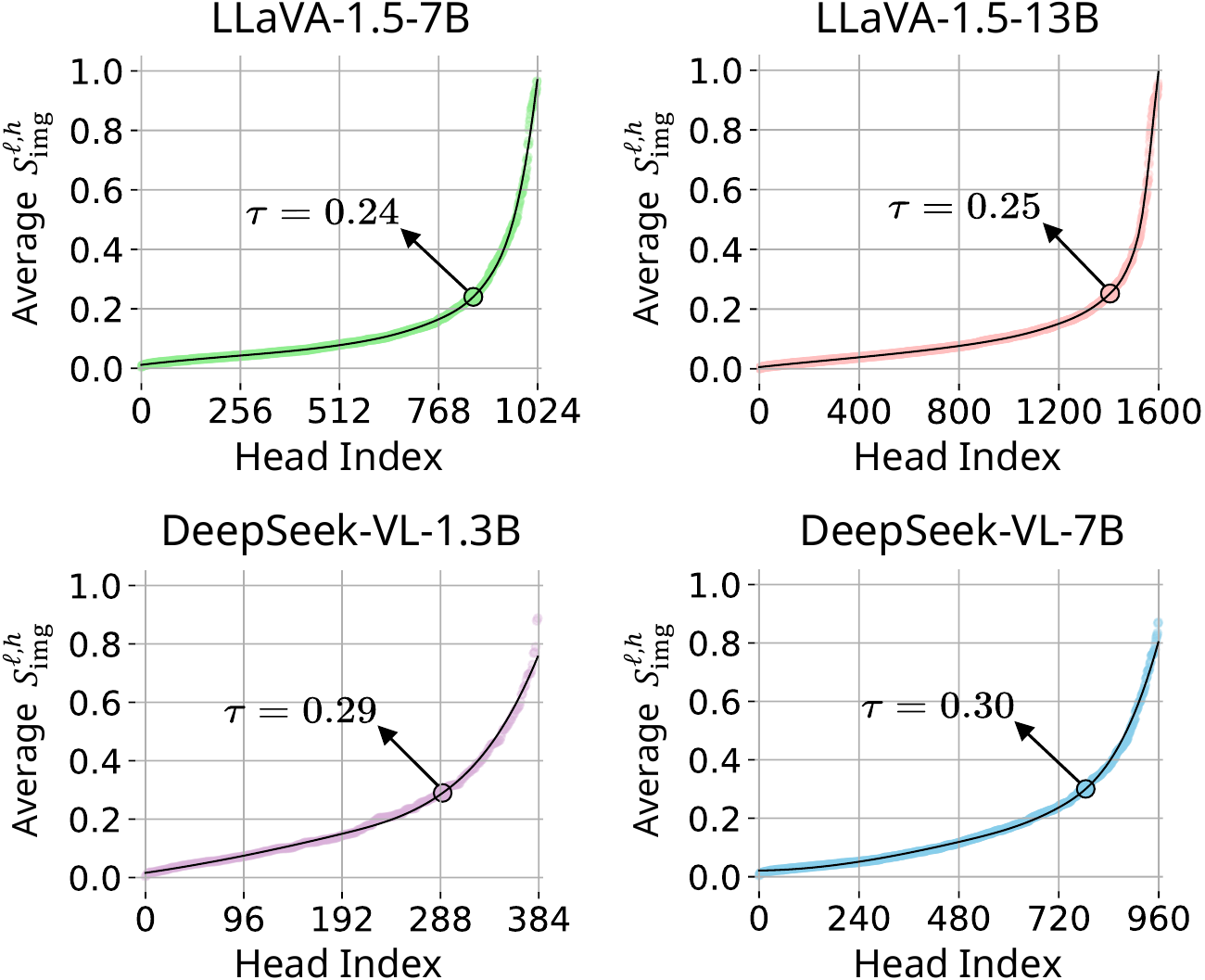}
    \vspace{-15pt}
    \caption{Average $S_{\text{img}}^{\ell,h}$ values for each attention head. We sort the heads in ascending order of $S_{\text{img}}^{\ell,h}$. Attention heads with $S_{\text{img}}^{\ell,h} \geq \tau$ are considered to effectively attend to the image, where $\tau$ is the threshold determined by the maximum curvature in the graph.}
    \vspace{-18pt}
    \label{fig:3}
\end{figure}

\subsection{Criteria to Find Localization Heads}
\label{sec:4_1}
Our final goal is to identify heads that excel in text referring. To achieve this, we propose two criteria in this section.

\vspace{3pt}
\noindent \textbf{Criterion 1: Attention Sum.}
To identify heads that predominantly focus on the overall image, we first consider attention sum $S_{\text{img}}^{\ell,h} = \sum_{i=1}^{P^2} \va^{\ell,h}[i]$, which quantifies the relevance of image information to $\vq_\text{txt}$ within individual attention heads. Then, the average $S_{\text{img}}^{\ell,h}$ for each head is computed across 1,000 random samples from RefCOCO~\cite{kazemzadeh2014referitgame} training set.

As shown in \cref{fig:3}, most attention heads exhibit low $S_{\text{img}}^{\ell,h}$ values, indicating that relatively few heads contribute significantly to the model's text-image interaction. To distinguish heads with high $S_{\text{img}}^{\ell,h}$ from those with low values, we set the threshold $\tau$ at the point of the maximum curvature in the graph (\eg, $\tau=0.24$ in LLaVA-1.5-7B~\cite{LLaVA1.5}). We deem the heads with $S_{\text{img}}^{\ell,h} \geq \tau$ to effectively attend to image. While we adopt the maximum curvature as a practical choice, we note that our analysis remains robust across a range of reasonable $\tau$ values. For analyses using alternative $\tau$ values, please refer to \appdx{Appendix~Sec.~C.}

\vspace{3pt}
\noindent \textbf{Criterion 2: Spatial Entropy.}
\label{sec:analyze_se}
For an attention head to be considered effective at focusing on objects, it must not only have a high attention sum value for the image but also concentrate its attention specifically around the objects. Since it is reasonable to assume that the object patches tend to stay near each other~\cite{simeoni2021lost, yun2022patch, wang2025sclip}, we evaluate how locally a cluster is formed in each attention map through spatial entropy~\cite{batty1974spatial, peruzzo2024spatialtransformer} to identify localization heads.

\begin{figure}[t]
    \centering
    \includegraphics[width=0.85\linewidth]{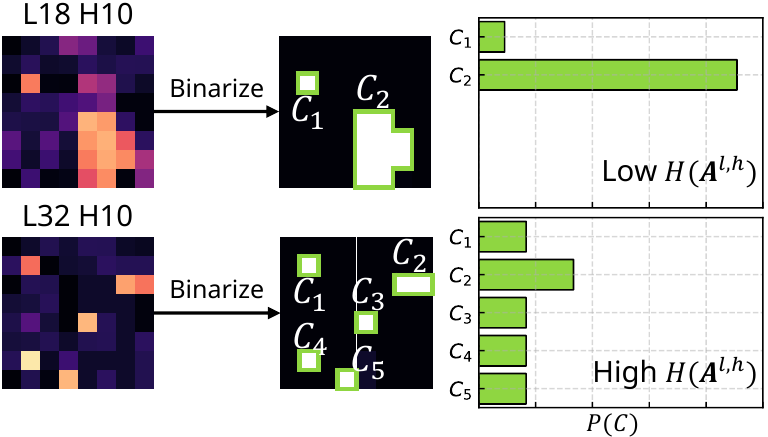}
    \vspace{-5pt}
    \caption{Illustration of the process for calculating spatial entropy. The attention map is binarized, and the spatial entropy is computed based on the sizes of its connected components $\{C_{i}\}_{i=1}^{N}$.}
    \label{fig:4}
\end{figure}

\cref{fig:4} presents an example of how spatial entropy is calculated. First, we reshape the attention weights $\va^{\ell,h}[1:P^2]$ into a $P \times P$ attention map $\mA^{\ell,h}$. The attention map is binarized by assigning a value of 1 to elements above the mean and 0 to those below it~\cite{peruzzo2024spatialtransformer}. Next, we identify connected components $C_i$~\cite{grana2010optimized}, defined as a set of coordinates connected via 8-neighbors. Then, for the set of $N$ connected components $\{C_{i}\}_{i=1}^{N}$, the spatial entropy $H$ is calculated as:
\begin{equation}
H(\mA^{\ell,h}) = -\sum_{i=1}^{N} P(C_i) \log{P(C_i)},
\end{equation}
where $P(C_i) = |C_i|/\sum_{i=1}^{N}|C_i|$. As a result, an attention map $\mA^{\ell,h}$ is considered effectively localized if it exhibits low spatial entropy. For more mathematical details on spatial entropy, please refer to the \appdx{Appendix~Sec.~B.} 

\begin{figure}[t]
    \centering
    \includegraphics[width=\linewidth]{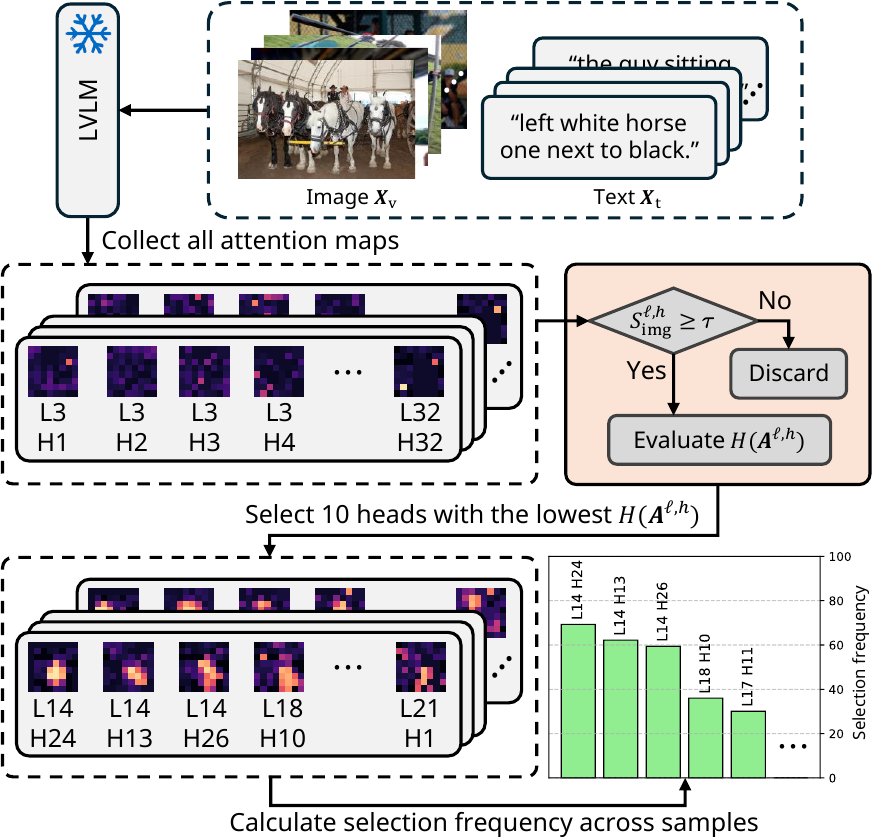}
    \vspace{-15pt}
    \caption{Overview of finding localization heads. We first identify heads with high attention sum. Then, we evaluate spatial entropy for each head and select 10 heads with the lowest spatial entropy. We repeat this process for 1,000 image-text pairs and calculate the selection frequency of each head.}
    \vspace{-12pt}
    \label{fig:5}
\end{figure}
\begin{figure*}[t]
    \centering
    \includegraphics[width=0.95\linewidth]{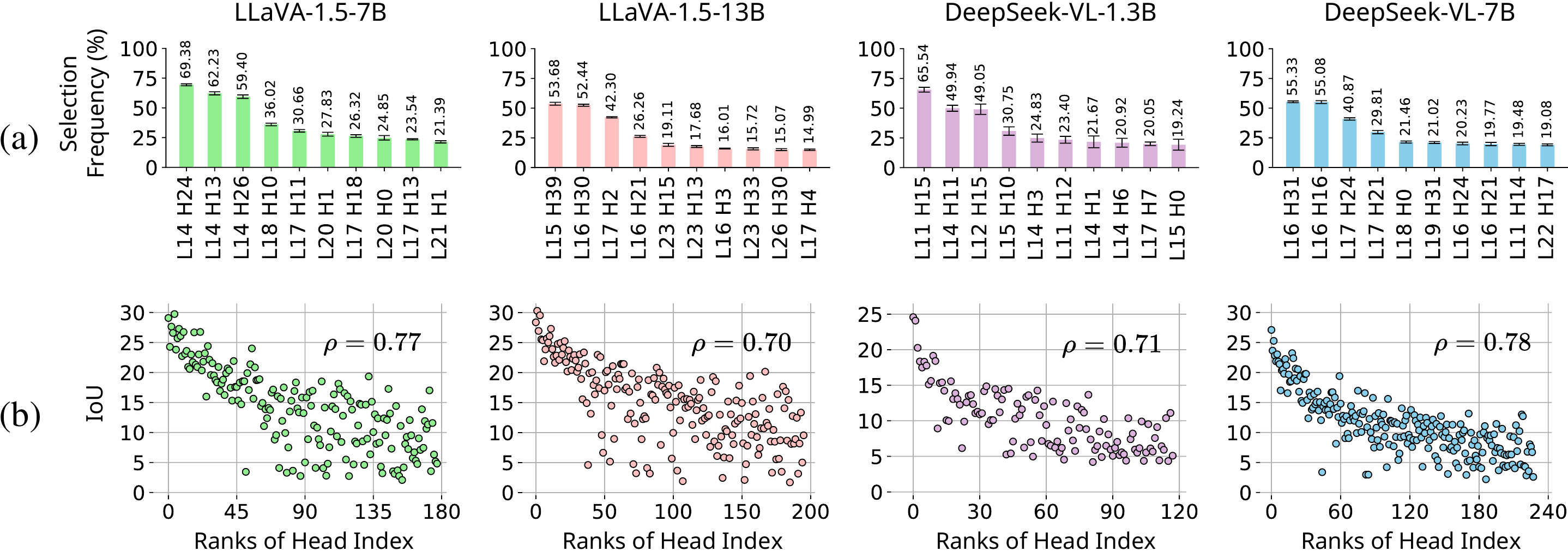}
    \vspace{-5pt}
    \caption{(a) Selection frequency of individual heads. Only a few heads exhibit high selection frequency, suggesting that their attention maps are consistently well-localized. We calculate the selection frequency five times and report the average and standard deviation. (b) Scatter plot illustrating the relationship between selection frequency rank and each head’s average IoU. Heads with higher selection frequency tend to show higher IoU values, indicating that they capture text semantics more effectively. The Spearman correlation coefficient ($\rho$) between rank and IoU is displayed in the top-right corner. The results of the Spearman correlation are statistically significant ($p < 0.001$).}
    \vspace{-7pt}
    \label{fig:6}
\end{figure*}
\subsection{Finding Localization Heads via Criteria}
\label{sec:4_2}

In this section, we utilize the two criteria described earlier to select a small subset of attention heads. Then, we demonstrate that the selected heads effectively capture objects relevant to the text.

To begin with, we rank all attention heads in order of how well they meet our criteria. Specifically, for 1,000 random image-text samples from the RefCOCO~\cite{kazemzadeh2014referitgame} training set, we retain all the heads that satisfy $S_{\text{img}}^{\ell, h} \geq \tau$. Among these heads, we calculate the frequency with which each head exhibits the 10-lowest spatial entropy across the samples to identify heads consistently exhibiting low spatial entropy. We refer to this metric as the selection frequency. The overall process is illustrated in \cref{fig:5}, and the results are reported in \cref{fig:6}(a). Now, we assign ranks to each head based on their selection frequency, with higher-ranked heads being those with high selection frequency. For example, in \cref{fig:6}(a), with LLaVA-1.5-7B~\cite{LLaVA1.5}, head L14 H24 ranks first, followed by head L14 H13 in second place.

Finally, we aim to demonstrate that higher-ranked heads are more effective at capturing objects relevant to the text. To this end, we binarize the attention maps of each head to obtain pseudo-masks and measure the IoU between these pseudo-masks and the ground truth (GT) masks. Then, we visualize the relationship between head ranks, derived from \cref{fig:6}(a), and their IoU values as a scatter plot, shown in \cref{fig:6}(b). Note that only the heads with a selection frequency of at least 1\% are considered in this analysis.

As visualized in \cref{fig:6}(b), attention heads with higher selection frequency tend to exhibit higher average IoU. We also calculate the Spearman correlation coefficient to quantitatively evaluate the relationship between the selection frequency and IoU. The correlation coefficients are above 0.7 for all LVLMs, indicating strong positive correlations.
This trend becomes increasingly evident for heads with higher ranks, leading us to conclude that a small number of top-ranked heads strongly capture semantic information. We refer to these heads as \textit{localization heads.} Since the trend consistently appears across various LVLMs (see \appdx{Appendix~Sec.~C.} for trends across more LVLMs), we claim that localization heads are an innate property of LVLMs.
\section{Visual Grounding with Localization Heads}
\label{sec:5}
\begin{figure}[t]
    \centering
    \includegraphics[width=\linewidth]{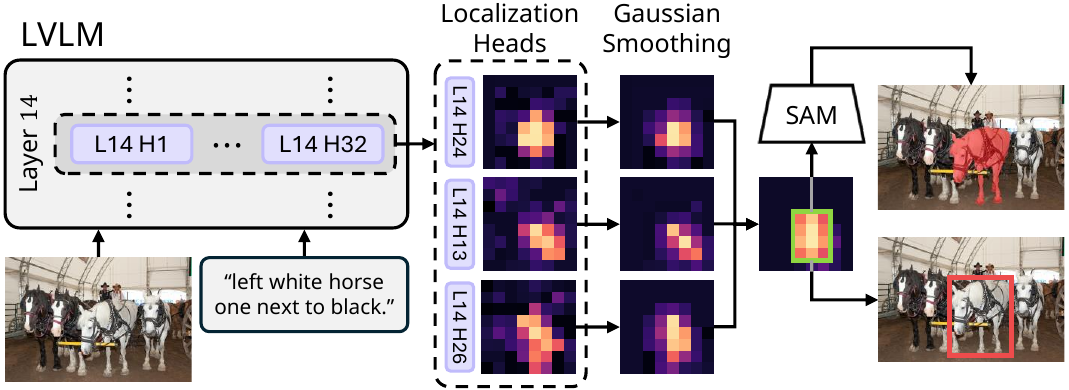}
    \vspace{-15pt}
    \caption{Our training-free visual grounding framework. Attention maps of localization heads are assembled into a combined map, which is then used to define the bounding box or segmentation mask.}
    \vspace{-17pt}
    \label{fig:7}
\end{figure}

In the previous section, we demonstrated that our criteria effectively identifies text-referring localization heads. Building on this, we propose a simple yet effective method to solve visual grounding tasks using these localization heads.

\begin{table*}[ht]

\setlength{\tabcolsep}{3.5pt}

\centering
\begin{minipage}{0.495\linewidth}

\caption{Comparison of our method with existing fine-tuning based and training-free methods on the REC (Referring Expression Comprehension) task. All fine-tuning based methods are trained on the training set of the corresponding datasets. Best performance is colored in \textcolor{red}{red} for fine-tuning and in \textcolor{blue}{blue} for training-free methods.}\label{tab:1}
\vspace{-5pt}

\resizebox{\linewidth}{!}{
\begin{tabular}{lcccccccc}
\toprule
\multicolumn{1}{c}{\multirow{2}{*}{Method}} & \multicolumn{3}{c}{RefCOCO} & \multicolumn{3}{c}{RefCOCO+} & \multicolumn{2}{c}{RefCOCOg} \\
\cmidrule(lr){2-4}
\cmidrule(lr){5-7}
\cmidrule(lr){8-9}
& val & testA & testB & val & testA & testB & val & test\\
\hline \hline

\rowcolor{gray!20}
\multicolumn{9}{l}{\textit{Fine-tuning based methods}} \\
MDETR~\cite{kamath2021mdetr} & 86.8 & 89.6 & 81.4 & 79.5 & 84.1 & 70.6 & 81.6 & 80.9 \\
SeqTR~\cite{zhu2022seqtr} & 87.0 & 90.2 & 83.6 & 78.7 & 84.5 & 71.9 & 82.7 & 83.4 \\
G-DINO~\cite{liu2023grounding} & 89.2 & 91.9 & 86.0 & 81.1 & 87.4 & 74.7 & 84.2 & 84.9 \\
ONE-PEACE~\cite{wang2023ONEPEACE} & 92.6 & 94.2 & 89.3 & \textcolor{red}{88.8} & 92.2 & 83.2 & 89.2 & 89.3 \\
UNINEXT~\cite{lin2023UNINEXT} & 92.6 & 94.3 & \textcolor{red}{91.5} & 85.2 & 89.6 & 79.8 & 88.7 & 89.4 \\ 

\hline

\rowcolor{gray!20}
\multicolumn{9}{l}{\textit{Fine-tuning based methods w/ LVLMs}} \\
Shikra-7B~\cite{chen2023shikra} & 87.0 & 90.6 & 80.2 & 81.6 & 87.4 & 72.1 & 82.3 & 82.2 \\
Ferret-7B~\cite{you2023ferret} & 87.5 & 91.4 & 82.5 & 80.8 & 87.4 & 73.1 & 83.9 & 84.8 \\
Shikra-13B~\cite{chen2023shikra} & 87.8 & 91.1 & 81.8 & 82.9 & 87.8 & 74.4 & 82.6 & 83.2 \\
Ferret-13B~\cite{you2023ferret} & 89.5 & 92.4 & 84.4 & 82.8 & 88.1 & 75.2 & 85.8 & 86.3 \\
CogVLM-17B~\cite{wang2023cogvlm} & \textcolor{red}{92.8} & \textcolor{red}{94.8} & 89.0 & 88.7 & \textcolor{red}{92.9} & \textcolor{red}{83.4} & \textcolor{red}{89.8} & \textcolor{red}{90.8} \\

\hline \hline

\rowcolor{gray!20}
\multicolumn{9}{l}{\textit{Training-free methods}} \\
ReCLIP~\cite{subramanian2022ReCLIP} & 45.8 & 46.1 & 47.1 & 47.9 & 50.1 & 45.1 & 59.3 & 59.0 \\
Han et al.~\cite{han2024triplet} & 49.4 & 47.8 & 51.7 & 48.9 & 50.0 & 46.9 & 61.0 & 60.0 \\ 
GroundVLP~\cite{shen2024groundvlp} & 65.0 & 73.5 & 55.0 & 68.8 & 78.1 & 57.3 & 74.7 & 75.0 \\
\hline

\rowcolor{gray!20}
\multicolumn{9}{l}{\textit{Training-free methods w/ LVLMs (Ours)}} \\
DeepSeek-VL-1.3B & 73.2 & 77.7 & 70.7 & 62.0   & 66.7 & 57.1 & 65.2 & 69.3 \\
Mini-Gemini-2B & 74.0 & 77.5 & 71.1 & 62.5 & 67.8 & 59.3 & 65.1 & 69.3 \\
InternVL-6B & 85.2 & 86.4 & 78.5 & 78.0 & 83.3 & 71.9 & 81.1 & 80.5 \\
Yi-VL-6B & 85.1 & 86.8 & 78.4 & 78.9 & 84.2 & 72.2 & 80.5 & 80.9 \\
DeepSeek-VL-7B & 85.3 & 87.2 & 81.0   & 77.8 & 83.9 & 73.5 & 81.1 & 82.8 \\
ShareGPT4V-7B & 86.1 & 87.1 & 80.5 & 79.7 & 86.2 & 71.3 & 82.4 & 82.9 \\ 
LLaVA-7B & 80.3 & 83.5 & 77.4 & 74.5 & 80.2 & 69.3 & 77.5 & 77.1 \\ 
LLaVA-1.5-7B & 86.5 & 89.8 & 80.2 & 80.1 & 86.3 & 71.9 & 82.3 & 83.0 \\
LLaVA-13B & 82.8 & 85.3 & 79.8 & 79.3 & 82.4 & 73.0   & 79.8 & 79.5 \\
LLaVA-1.5-13B & \textcolor{blue}{87.2} & \textcolor{blue}{90.0} & \textcolor{blue}{83.3} & \textcolor{blue}{82.7} & \textcolor{blue}{88.5} & \textcolor{blue}{74.0} & \textcolor{blue}{84.3} & \textcolor{blue}{85.5} \\
  
\bottomrule

\end{tabular}
}

\end{minipage}
\hfill
\begin{minipage}{0.495\linewidth}

\setlength{\tabcolsep}{3.5pt}

\centering

\caption{Comparison of our method with existing fine-tuning based and training-free methods on the RES (Referring Expression Segmentation) task. All fine-tuning based methods, except for LISA~\cite{lai2024lisa} and GSVA~\cite{xia2024gsva}, are trained on the training set of the corresponding datasets. \textcolor{red}{Red} and \textcolor{blue}{blue} colors are used as in \cref{tab:1}.}\label{tab:2}
\vspace{-5pt}

\resizebox{\linewidth}{!}{
\begin{tabular}{lcccccccc}
\toprule
\multicolumn{1}{c}{\multirow{2}{*}{Method}} & \multicolumn{3}{c}{RefCOCO} & \multicolumn{3}{c}{RefCOCO+} & \multicolumn{2}{c}{RefCOCOg} \\
\cmidrule(lr){2-4}
\cmidrule(lr){5-7}
\cmidrule(lr){8-9}
& val & testA & testB & val & testA & testB & val & test\\

\hline \hline

\rowcolor{gray!20}
\multicolumn{9}{l}{\textit{Fine-tuning based methods}} \\
LAVT~\cite{yang2022lavt} & 72.7 & 75.8 & 68.8 & 62.1 & 68.4 & 55.1 & 61.2 & 62.1 \\
ReLA~\cite{liu2023gres} & 73.8 & 76.5 & 70.2 & 66.0 & 71.0 & 57.7 & 65.0 & 66.0 \\
UniRef++~\cite{wu2023uniref++} & 79.1 & 82.1 & 77.5 & 68.4 & 74.0 & 61.5 & 71.4 & 72.8 \\
UNINEXT~\cite{lin2023UNINEXT} & 82.2 & 83.4 & 81.3 & 72.5 & 76.4 & 66.2 & 74.4 & 76.4 \\ 

\hline

\rowcolor{gray!20}
\multicolumn{9}{l}{\textit{Fine-tuning based methods w/ LVLMs}} \\
LISA-7B~\cite{lai2024lisa} & 74.1 & 76.5 & 71.1 & 62.4 & 67.4 & 56.5 & 66.4 & 68.5 \\
GSVA-7B~\cite{xia2024gsva} & 76.4 & 77.4 & 72.8 & 64.5 & 67.7 & 58.6 & 71.1 & 72.0 \\
LISA-13B~\cite{xia2024gsva} & 73.4 & 76.2 & 69.5 & 62.3 & 66.6 & 56.3 & 68.2 & 68.5 \\
GSVA-13B~\cite{xia2024gsva} & 77.7 & 79.9 & 74.2 & 68.0 & 71.5 & 61.5 & 73.2 & 73.9 \\
GLaMM~\cite{rasheed2024glamm} & 79.5 & 83.2 & 76.9 & \textcolor{red}{75.9} & \textcolor{red}{78.7} & 68.8 & \textcolor{red}{76.8} & \textcolor{red}{78.4} \\
PSALM~\cite{zhang2024psalm} & \textcolor{red}{83.6} & \textcolor{red}{84.7} & \textcolor{red}{81.6} & 72.9 & 75.5 & \textcolor{red}{70.1} & 73.8 & 74.4 \\

\hline \hline

\rowcolor{gray!20}
\multicolumn{9}{l}{\textit{Training-free methods}} \\
Yu et al.~\cite{yu2023ZSRS} & 24.9 & 23.6 & 24.7 & 26.2 & 24.9 & 25.8 & 31.1 & 31.0 \\ 
TAS~\cite{suo2023TAS} & 29.5 & 30.3 & 28.2 & 33.2 & 38.8 & 28.0 & 35.8 & 36.2 \\ 
Ref-Diff~\cite{ni2023RefDiff} & 35.2 & 37.4 & 34.5 & 35.6 & 38.7 & 31.4 & 38.6 & 37.5 \\ 

\hline

\rowcolor{gray!20}
\multicolumn{9}{l}{\textit{Training-free methods w/ LVLMs (Ours)}} \\
DeepSeek-VL-1.3B & 56.3 & 57.0 & 52.7 & 51.2 & 55.5 & 49.2 & 52.3 & 55.8 \\ 
Mini-Gemini-2B & 59.8 & 60.3 & 55.5 & 56.3 & 59.9 & 51.8 & 55.1 & 60.3 \\
InternVL-6B & 62.1 & 65.8 & 60.9 & 62.2 & 65.5 & 55.5 & 63.5 & 65.4 \\
Yi-VL-6B & 62.5 & 65.8 & 60.7 & 61.0 & 65.3 & 56.0 & 64.0 & 67.0 \\
DeepSeek-VL-7B & 73.9 & 76.6 & 70.7 & 63.1 & 66.1 & 56.5 & 64.0 & 68.9 \\ 
ShareGPT4V-7B & 73.5 & 76.7 & 70.1 & 59.4 & 63.8 & 55.9 & 60.7 & 65.1 \\ 
LLaVA-7B & 65.4 & 66.2 & 61.1 & 59.9 & 63.2 & 52.7 & 59.7 & 63.3 \\ 
LLaVA-1.5-7B & 74.2 & 76.5 & 70.4 & 62.5 & 65.2 & 56.0 & 64.2 & 68.1 \\ 
LLaVA-13B & 66.8 & 68.0 & 63.7 & 62.3 & 66.9 & \textcolor{blue}{57.3} & 65.0 & 68.2 \\ 
LLaVA-1.5-13B & \textcolor{blue}{76.1} & \textcolor{blue}{78.9} & \textcolor{blue}{72.8} & \textcolor{blue}{64.1} & \textcolor{blue}{67.1} & \textcolor{blue}{57.3} & \textcolor{blue}{67.7} & \textcolor{blue}{69.0} \\
\bottomrule

\end{tabular}
}

\end{minipage}

\vspace{-8pt}
\end{table*}

Specifically, our objective is to perform visual grounding tasks, given an LVLM. To achieve this, the localization heads of the LVLM must first be identified. Following the process we described in \cref{sec:4_2} and \cref{fig:5}, we rank the heads based on the selection frequency and select the heads with the $k$-highest rank. Subsequently, an image-text pair for which a mask is to be generated is fed into the LVLM, and attention maps are extracted from the localization heads.

As illustrated in \cref{fig:7}, Gaussian smoothing is applied to each attention map of the localization head to preserve detailed localization information while minimizing potential random noise. The resulting maps are assembled through element-wise summation to produce the combined map. This combined map is then binarized to produce the pseudo-mask. Finally, the largest rectangle encompassing the pseudo-mask is identified and can be used as a bounding box. Additionally, this bounding box can serve as a prompt for SAM~\cite{SAM} to address the segmentation task. Additional details on the algorithm used to find the bounding box are provided in \appdx{Appendix.~Sec.~B}, and the ablation study on Gaussian smoothing is presented in \appdx{Appendix.~Sec.~D}.
\section{Experiments}
\label{sec:6}
\begin{figure*}[t!]
    \centering
    \includegraphics[width=\linewidth]{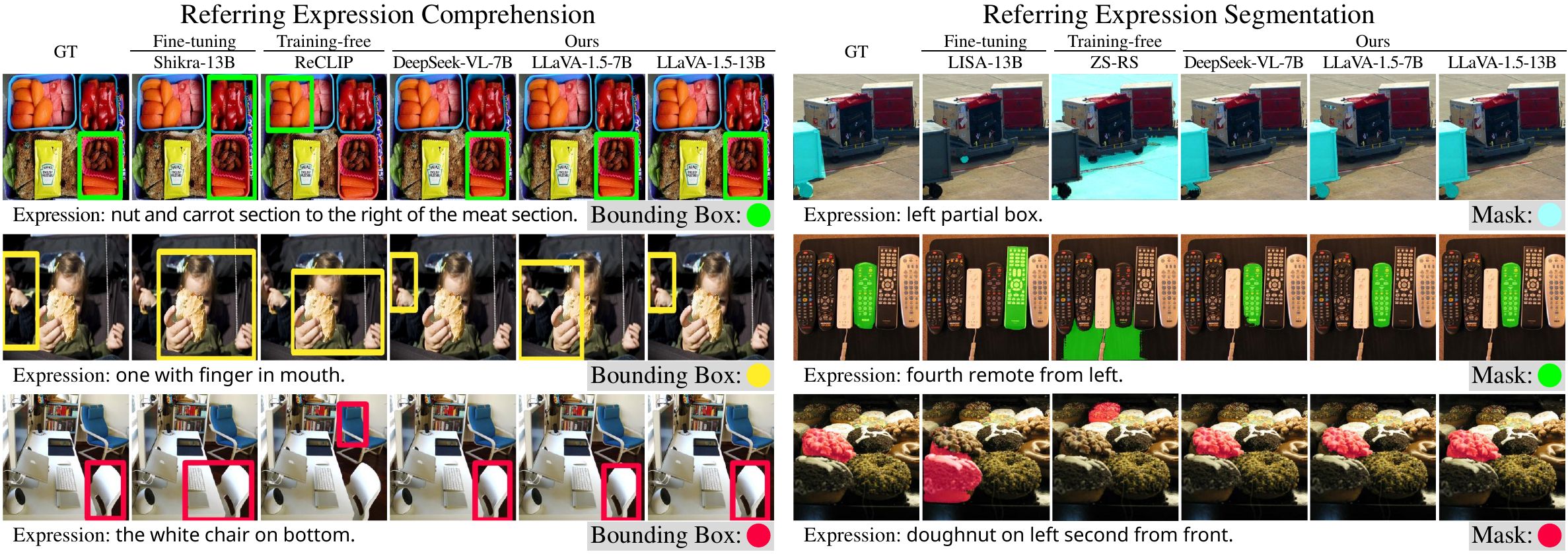}
    \vspace{-18pt}
    \caption{Qualitative results of our framework with the baseline models. LVLMs successfully localize the referred objects in various challenging scenarios including multiple similar objects, non-salient objects, and complex spatial relations.}
    \vspace{-8pt}
    \label{fig:8}
\end{figure*}

In this section, we verify whether the localization head discovered through our selection process ensures robust performance on well-known visual grounding benchmarks. Additionally, we conduct ablation studies to validate the settings of our method.

\subsection{Experimental Setup}
\noindent\textbf{Models.} We apply our approach across ten LVLMs to validate its broad applicability. The main experiments include DeepSeek-VL~\cite{lu2024deepseek}, Mini-Gemini~\cite{li2024mini-gemini}, InternVL~\cite{chen2024internvl}, Yi-VL~\cite{young2024yi}, ShareGPT4V~\cite{chen2023sharegpt4v}, LLaVA~\cite{LLaVA}, and LLaVA-1.5~\cite{LLaVA1.5}, with model sizes ranging from 1.3B to 13B. The number of localization heads is fixed to $k=3$ for all models.

\noindent\textbf{Benchmarks.} To assess visual grounding capabilities, we conduct experiments on Referring Expression Comprehension (REC) and Referring Expression Segmentation (RES) tasks. REC requires the model to predict the bounding box of the referred object, while RES requires the segmentation mask. We use the RefCOCO, RefCOCO+~\cite{kazemzadeh2014referitgame}, and RefCOCOg~\cite{hu2016res1} datasets. We further evaluate the performance of our method on the more challenging scenario, Reasoning Segmentation (ReasonSeg)~\cite{lai2024lisa}, which requires complex reasoning or world knowledge. For the REC task, we report the performance using Acc@0.5 metric, which is the standard detection metric for REC. For the RES and ReasonSeg task, cIoU is used as the evaluation metric.

\noindent\textbf{Baselines.} We compare our method with existing fine-tuning based and training-free approaches. The fine-tuning based methods include visual grounding specialist models~\cite{kamath2021mdetr, zhu2022seqtr, liu2023grounding, lin2023UNINEXT, yang2022lavt, liu2023gres, wu2023uniref++}, along with fine-tuned LVLMs for object localization~\cite{chen2023shikra, you2023ferret, wang2023cogvlm} or segmentation tasks~\cite{lai2024lisa, xia2024gsva, rasheed2024glamm, zhang2024psalm}. The training-free methods include CLIP-based methods~\cite{subramanian2022ReCLIP, han2024triplet, yu2023ZSRS, suo2023TAS, shen2024groundvlp} and DM-based method~\cite{ni2023RefDiff}. More details on the experimental setup are provided in the \appdx{Appendix~Sec.~A.}

\subsection{Main Results}
\label{sec:main-results}

\noindent\textbf{REC and RES.} \cref{tab:1} and \cref{tab:2} present the results of our method and the baseline models on the REC and RES tasks, respectively. Our framework achieves substantial improvements over the existing training-free methods. Surprisingly, our method performs comparably to the fine-tuned LVLMs, even though our method does not require additional training. For example, in the REC task, the best performance of our approach achieves results on par with Shikra~\cite{chen2023shikra} and Ferret~\cite{you2023ferret}, which share the same base LLMs as LLaVA-1.5~\cite{LLaVA1.5}, but are fine-tuned for localization tasks. A similar finding is observed with LISA~\cite{lai2024lisa} in the RES task. The results indicate that frozen LVLMs can effectively localize the referred object without any additional training, due to the presence of localization heads.

\begin{table}[t]
    \centering
    \caption{Comparison of our framework with LISA~\cite{lai2024lisa} on the ReasonSeg (Reasoning Segmentation) benchmark.}\label{tab:3}
    \vspace{-5pt}
    \resizebox{.94\linewidth}{!}{
    \begin{tabular}{lccccc}
        \toprule
        \multicolumn{1}{c}{\multirow{2}{*}{Method}} & val & \multicolumn{3}{c}{test} \\
        \cmidrule(lr){2-2}
        \cmidrule(lr){3-5}
        & overall & short & long & overall \\
        \hline
        LISA-7B~\cite{lai2024lisa}       & 52.3 & 48.5 & 48.9 & 48.8  \\
        LISA-13B~\cite{lai2024lisa}      & 60.3 & 50.0 & 50.9 & 50.8  \\ \hline
        LLaVA-1.5-7B (Ours)  & 52.4 & 48.0 & 49.1 & 48.7  \\
        LLaVA-1.5-13B (Ours) & 60.5 & 48.7 & 51.0 & 49.9  \\
        \bottomrule
    \end{tabular}
    }
\vspace{-10pt}
\end{table}

Notably, the visual grounding capability is enhanced as the model evolves. First, performance consistently improves as model size increases (1.3B to 13B). Second, updates in architecture and training data (\eg, LLaVA to LLaVA-1.5) also boost performance. This observation suggests that the grounding ability of LVLMs could be further enhanced with larger models and more diverse training data.

\cref{fig:8} compares the qualitative results of our method with those of the baseline models. The results demonstrate that LVLMs can accurately identify the correct object regions, even in challenging scenarios where multiple similar objects are present, or when the referred object is not prominently centered in the image. According to \cite{subramanian2022ReCLIP}, CLIP-based methods struggle to interpret orientation descriptors (\eg, ``left''). Therefore, they have to manually decompose the referring expression into multiple components~\cite{yu2023ZSRS} or rely on post-processing steps that use the object's spatial information~\cite{suo2023TAS}. In contrast, our framework can directly predict the bounding box or segmentation mask of the referred object without carefully designed post-processing steps, with the help of the strong text comprehension capabilities of LVLMs. More qualitative results are provided in the \appdx{Appendix~Sec.~E.}

\vspace{3pt}
\noindent\textbf{Reasoning Segmentation.} \cref{tab:3} shows the results of our method and LISA~\cite{lai2024lisa} on the ReasonSeg. For a fair comparison, we compare both methods using the same backbone model, LLaVA-1.5~\cite{LLaVA1.5}. Our method performs comparably to LISA and sometimes outperforms it. The results suggest that the localization heads in LVLMs are generalizable to various visual grounding tasks, including those that require complex reasoning or world knowledge.

\subsection{Ablation Studies}

\noindent\textbf{Number of Localization Heads.} In our main experiments, we set the number of localization heads to $k=3$. Here, we investigate the effect of varying $k$ on visual grounding performance. \cref{tab:4} presents the results of our framework with different $k$ values. We observe that the performance generally improves as $k$ increases from 1 to 3, indicating that top-3 heads complement each other to provide more accurate localization. However, increasing $k$ further does not guarantee better performance, implying that additional heads may introduce noise or redundancy. It is worth noting that the optimal $k$ trend remains consistent across different LVLMs. The results suggest that similar numbers of attention heads are responsible for localization of referred objects in various LVLMs, even though the total number of heads and model architectures differ.

\begin{table}[t]
    \centering
    \caption{Ablation study on the number of localization heads ($k$) on the RefCOCO validation set for the RES task.}\label{tab:4}
    \vspace{-5pt}
    \resizebox{.92\linewidth}{!}{
    \begin{tabular}{lccccc}
        \toprule
        \multicolumn{1}{c}{\multirow{2}{*}{Method}} & \multicolumn{5}{c}{$k$ (\# of Localization Heads)} \\
        \cmidrule(lr){2-6}
        & 1 & 2 & 3 & 4 & 5 \\
        \hline
        DeepSeek-VL-1.3B & 55.1 & {56.3} & {56.3} & 55.3 & 51.2\\
        MiniGemini-2B	& 58.0	& 58.5	& {59.8}	& 59.1	& 54.2 \\
        InternVL-6B		& 61.3	& 61.8	& {62.1}	& 61.0	& 55.7 \\
        Yi-VL-6B         & 61.8 & 62.1 & 62.5 & {62.6} & 55.4\\
        DeepSeek-VL-7B   & 70.1 & 72.2 & {73.9} & 73.0 & 65.3\\
        ShareGPT4V-7B    & 70.3 & 72.4 & {73.5} & {73.5} & 60.8\\
        LLaVA-7B         & 62.7 & 63.1 & {65.4} & 65.3 & 57.7\\
        LLaVA-1.5-7B     & 70.3 & 73.1 & {74.2} & 74.1 & 65.4\\
        LLaVA-13B        & 63.5 & 64.7 & {66.8} & 66.4 & 57.8\\
        LLaVA-1.5-13B    & 71.7 & 75.7 & {76.1} & 76.0 & 65.7\\
        \hline
        Average & 64.5 & 66.0 & \textcolor{blue}{67.1} & 65.4 & 58.9  \\
        \bottomrule
    \end{tabular}
    }
\vspace{-5pt}
\end{table}
\begin{table}[t]
    \centering
    \caption{Ablation study on the validation of criteria and selection methods for localization heads. The results are reported on the RefCOCO validation set and LLaVA-1.5-13B.}\label{tab:5}
    \vspace{-5pt}
    \resizebox{.84\linewidth}{!}{
    \begin{tabular}{cccccc}
        \toprule
        \multicolumn{2}{c}{Criteria} & \multicolumn{2}{c}{Selection} & \multirow{2}{*}{REC} & \multirow{2}{*}{RES} \\
        \cmidrule(lr){1-2} \cmidrule(lr){3-4}
        $S^{\ell, h}_{\text{img}}$ & $H(\mA^{\ell,h})$ & Fixed & Greedy & \\
        \hline
        \cmark & & & \cmark & 23.7 & 18.8 \\
        & \cmark & & \cmark & 29.8 & 21.5 \\
        \cmark & \cmark & & \cmark & 67.4 & 63.8 \\
        \hline
        \cmark & & \cmark & & 23.9 & 19.3 \\
        & \cmark & \cmark & & 31.3 & 25.7 \\
        \cmark & \cmark & \cmark & & \textcolor{blue}{87.2} & \textcolor{blue}{76.1} \\
        \bottomrule
    \end{tabular}
    }
\vspace{-12pt}
\end{table}

\vspace{3pt}
\noindent\textbf{Validation of Criteria and Selection Methods for Localization Heads.}
In \cref{sec:4_1}, we propose two criteria, attention sum $S^{\ell, h}_{\text{img}}$ and spatial entropy $H(\mA^{\ell,h})$, to identify localization heads. Then, we select the fixed top-$k$ heads based on the selection frequency, as described in \cref{sec:4_2}. We ablate the effectiveness of each criterion and validate selection methods. For criterion ablation, we evaluate the performance of our method using each criterion individually: (1) selecting heads with either the highest $S^{\ell, h}_{\text{img}}$ or (2) the lowest $H(\mA^{\ell,h})$ only. For selection validation, we compare the performance of our method (denoted as the `fixed' method for comparison) with `greedy' selection, where the top-$k$ heads are selected and aggregated per sample. Further details regarding the settings are provided in \appdx{Appendix~Sec.~A.}

\cref{tab:5} shows the results of these ablation studies.
The performance drops significantly when only one criterion is used, indicating that both criteria are essential for identifying localization heads. Furthermore, the greedy selection method shows worse results than the fixed method. While our criteria identify attention maps exhibiting apparent clusters, they do not ensure that these clusters are formed around text semantics. As a result, the greedy method may select heads that are localized but not text-referred. In contrast, our method involves a statistical analysis (\ie, selection frequency). This ensures that the localization heads are genuinely text-referred, consistently focusing on text-related regions rather than arbitrarily clustered areas.

\begin{figure}[t]
    \centering
    \includegraphics[width=.9\linewidth]{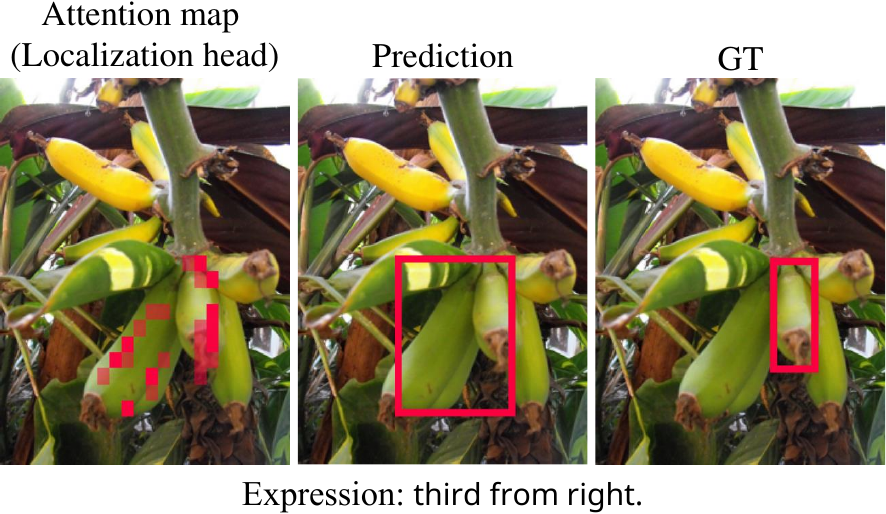}
    \vspace{-7pt}
    \caption{Failure case of the LLaVA-1.5-13B~\cite{LLaVA1.5} in visual grounding. The text-to-image attention map from a localization head (L15 H39) shows where the model focuses, helping to understand the model's failure.}
    \vspace{-12pt}
    \label{fig:9}
\end{figure}

\subsection{Understanding LVLMs When They Fail}

Here, we briefly discuss how localization heads may also help us better understand LVLMs. Specifically, localization heads allow us to identify where LVLMs focus when they fail to ground the correct object. \cref{fig:9} illustrates an example where the model fails to predict the correct object, the third \texttt{banana} from the right. As shown in the first column of \cref{fig:9}, the text-to-image attention map from a localization head focuses on both the third and fourth \texttt{bananas} from the right. This observation suggests that LVLMs struggle with pinpointing the exact location of objects.
These findings show the localization head's potential to provide a transparent understanding of where the LVLMs focus.
\section{Conclusion}
\label{sec:7}
In this work, we identify \textit{localization heads} within various LVLMs via criteria, which exhibit strong visual grounding capabilities in response to textual queries. We then propose a simple yet effective training-free framework that assembles the text-to-image attention maps from a few localization heads to predict bounding boxes and segmentation masks for text-relevant regions in the image. Our approach achieves competitive performance compared to fine-tuning based methods. Therefore, we conclude that LVLMs can act as text-referring localizers for visual grounding tasks with their inherent property under the attention mechanisms. We hope that our work opens up new possibilities for analyzing and utilizing the attention mechanisms of LVLMs.
\newpage\clearpage
{
    \small
    \bibliographystyle{ieeenat_fullname}
    \bibliography{main}
}

\newpage\clearpage
\begingroup
\setcounter{section}{0}
\renewcommand{\thesection}{\Alph{section}}
\crefname{section}{Sec.}{Secs.}
\Crefname{section}{Section}{Sections}
\Crefname{table}{Table}{Tables}
\crefname{table}{Tab.}{Tabs.}

\addcontentsline{toc}{section}{Appendix}
\begin{center}
    \textbf{\LARGE Appendix}
\end{center}
\section{Experimental Details}
\noindent\textbf{Experimental Setting.} All experiments and evaluations are conducted on a single NVIDIA GeForce RTX A6000 48GB GPU. We only use the inference stage of the models without any fine-tuning or training.

\noindent\textbf{Analysis Setting.} In Sec.~4, we identify and analyze localization heads in various LVLMs. We use the RefCOCO training set to prevent validation set leakage. To calculate the selection frequency of individual heads, we randomly select 1,000 image-text pair samples from the RefCOCO training set and average the results over five trials to validate consistency. When analyzing the selection frequency and IoU, we binarize the attention weights by assigning 1 above the mean value and 0 below it and calculate the IoU between the binarized attention weights and the ground-truth mask. We repeat this process for 1,000 image-text pairs and average the IoU scores.

\noindent\textbf{Dataset Details.} We evaluate our method on the following datasets:
\begin{itemize}
    \item RefCOCO, RefCOCO+~\cite{kazemzadeh2014referitgame}, and RefCOCOg~\cite{hu2016res1} datasets, sourced from MS-COCO~\cite{lin2014microsoft}, offer a collection of referring expressions and associated images. RefCOCO consists of 19,994 images paired with 142,210 expressions, while RefCOCO+ includes 19,992 images and 141,564 expressions. RefCOCOg, on the other hand, contains 26,771 images and 104,560 expressions. The expressions in RefCOCO and RefCOCO+ are generally concise, with an average of 1.6 nouns and 3.6 words per expression. In contrast, RefCOCOg features more descriptive expressions, averaging 2.8 nouns and 8.4 words.
    \item ReasonSeg: The dataset and benchmark for reasoning segmentation were first introduced in LISA~\cite{lai2024lisa}. The resulting ReasonSeg benchmark consists of 1,218 image-instruction-mask data samples, which are further divided into three splits: training (239 samples), validation (200 samples), and test (779 samples).
\end{itemize}

\noindent\textbf{Main Experiments Setting.} We evaluate our method on the following tasks: 
\begin{itemize}
    \item Referring Expression Comprehension (REC) and Referring Expression Segmentation (RES): The datasets evaluated for the main results in Sec.6.2 include RefCOCO (validation, test-A, test-B), RefCOCO+ (validation, test-A, test-B), and RefCOCOg (validation, test). All evaluations were conducted using the UNC split.
    \item Reasoning Segmentation (ReasonSeg): Reasoning Segmentation was first introduced in LISA~\cite{lai2024lisa}. This task shares a similar formulation with the referring expression segmentation task but is considerably more challenging. The key distinction lies in the complexity of the query text in reasoning segmentation. Rather than simple phrases (e.g., ``the blue mug''), the queries involve more nuanced descriptions (e.g., ``the container used for drinking, located next to the plate") or longer sentences (e.g., ``Find the item on the table that someone would use to hold liquid, often paired with a saucer''). These queries demand advanced reasoning and a deeper understanding of contextual and world knowledge. All reasoning segmentation results were evaluated using the ReasonSeg benchmark, which includes both the validation set and test set. Performance was measured across short queries, long queries, and overall, following the same experimental setup as LISA~\cite{lai2024lisa} to ensure consistency in comparisons.
\end{itemize}

\noindent\textbf{Ablation Studies Setting.} In Sec.~6.3, we ablate the effectiveness of each criterion and validate the selection methods. In this section, we provide details of the ablation studies.

For criterion ablation, we consider two approaches: (1) selecting heads based solely on the highest $S^{\ell, h}_{\text{img}}$ values, or (2) selecting heads based solely on the lowest $H(\mA^{\ell,h})$ values. In approach (1), we select the 10 heads with the highest $S^{\ell, h}_{\text{img}}$ values and calculate their selection frequency. Similarly, in approach (2), we select the 10 heads with the lowest $H(\mA^{\ell,h})$ values and calculate their selection frequency.

For selection validation, we introduce the `greedy' selection method, which selects the top-$k$ heads per sample without considering the overall selection frequency. When applying the greedy selection method and criterion (1) simultaneously, we select the top-$k$ heads with the highest $S^{\ell, h}_{\text{img}}$ values for each sample. Criterion (2) is applied in a similar manner, simultaneously selecting the top-$k$ heads with the lowest $H(\mA^{\ell,h})$ values for each sample.

\section{Detailed Description of Algorithms}
\subsection{Spatial Entropy}

Spatial entropy~\cite{batty1974spatial} adjusts the probability of attention being focused in a region by factoring in the size of that region, ensuring fair comparison across areas of different sizes. Note that, our spatial entropy calculation is based on the previous work~\cite{peruzzo2024spatialtransformer} which validated the effectiveness of spatial entropy in image attention maps within vision transformer.
We begin by computing the image attention map $ \mA^{\ell, h} $ as follows:
\begin{equation}
    \mA^{\ell, h} = \text{ReLU}\left(\text{reshape}(\va^{\ell, h}) - m\right),
\end{equation}
where the ReLU function is applied after reshaping by $P \times P$, and it retains only those values in $ \va^{\ell, h} $ that are greater than the mean $ m $. Next, we identify the connected components $ C_{\mA^{\ell, h}} = \{C_{1}, C_{2}, \dots, C_{n}\} $ from $ \mA^{\ell, h} $:
\begin{equation}
C_{\mA^{\ell, h}} = \text{ConnectedComponents}(\mA^{\ell, h}),
\end{equation}
where the connected components are determined by applying an 8-connectivity relation among the non-zero elements of $ \mA^{\ell, h} $. Each connected component $ C_{n} $ (with $ 1 \leq n \leq N $) in $ C_{\mA^{\ell, h}} $ is defined as the set of coordinates $C_n = \{(x_1, y_1), (x_2, y_2), \dots, (x_{k_n}, y_{k_n})\}$ for the $ n $-th component, where $ k_n = \vert C_n \vert $ represents the cardinality, or the number of elements, in $ C_n $. Finally, we calculate the spatial entropy $ H(\mA^{\ell,h}) $ as follows:
\begin{equation}
H(\mA^{\ell,h}) = -\sum_{n=1}^{N} P(C_n) \log{P(C_n)},
\end{equation}
where this entropy is computed using Shannon's entropy formula. Here, $ P(C_n) $ represents the probability of observing each connected component $ C_n $ within $ \mA^{\ell, h} $. The probability $ P(C_n) $ for each component $ C_n $ is defined as:
\begin{equation}
P(C_n) = \frac{|C_n|}{\sum_{n=1}^{N}|C_n|},
\end{equation}
where $ P(C_n) $ is calculated by dividing the area of $ C_n $ by the total area of all components in $ \mA^{\ell, h} $. This provides a normalized measure of spatial focus. The resulting spatial entropy $ H(\mA^{\ell,h}) $ ranges from 0 to 1. A value of 0 indicates that attention is completely focused on a single region, while a value of 1 suggests that attention is evenly distributed across the image. This measure thus enables us to evaluate the dispersion of the model's attention across different regions within the image. 

\subsection{Details of Our Framework}

In this section, we provide a detailed description of our framework, described in Sec.~5 of the main paper.

\noindent\textbf{Binarization of the Attention Map.} The attention map is binarized by setting values above the mean to 1. This approach effectively highlights the most significant regions of the attention map.

\noindent\textbf{Gaussian Smoothing.} Gaussian smoothing is applied using a kernel size of $k = 7$ and a standard deviation of $\sigma = 1.0$. These parameters ensure a balance between smoothing effects and detail preservation.

\noindent\textbf{Convex Hull Algorithm for Bounding Box.} To determine the bounding box in an assembled attention map from the localization heads, we employ the convex hull algorithm~\cite{graham1972efficient}. In cases where multiple convex hulls are present within the same attention map, we retain only the largest convex hull. Subsequently, we calculate the smallest tight bounding box that encloses the retained convex hull and we use it as  the final bounding box.

\section{More Analysis on Localization Heads}
\subsection{Extended Analysis Across More LVLMs}

In this section, we extend the analysis of localization heads in Sec.~\ref{sec:4} of the main paper to more LVLMs, including InternVL~\cite{chen2024internvl}, LLaVA~\cite{LLaVA}, Mini-Gemini~\cite{li2024mini-gemini}, ShareGPT4V~\cite{chen2023sharegpt4v}, and Yi-VL~\cite{young2024yi}.

\noindent\textbf{Average Attention Sum in More LVLMs.} We extend Fig.~3 in the main paper to demonstrate that relatively few attention heads significantly contribute to the model's text-image interaction. As shown in \cref{fig:11}, the trend of the average $S_{\text{img}}^{\ell, h}$ values remains consistent across different LVLMs.

\noindent\textbf{Selection Frequency and IoU in More LVLMs.} Similar to the above, we extend Fig.~6 in the main paper to cover additional LVLMs. \cref{fig:12} presents the selection frequency and a scatter plot of selection frequency rank versus IoU for each attention head across various LVLMs. The results confirm that our observations hold consistently across different LVLMs.

\subsection{Robustness of Localization Head Selection}

In this section, we validate the robustness of our localization head selection method across different threshold values ($\tau$) and the number of selected heads ($N$). The experiments below indicate that localization head selection is not sensitive to the choice of $\tau$ or $N$.

\noindent\textbf{Threshold $\tau$.} \cref{fig:3} in the main paper presents the average $S_{\text{img}}$ values for each attention head, setting the threshold $\tau$ at the point where the maximum curvature is observed. We select maximum curvature as the threshold to reduce the need for manual tuning; however, other $\tau$ values can also be considered. Therefore, we further validate that plausible $\tau$ values can give consistent results with the maximum curvature. To this end, we calculate the selection frequency of the heads based on different $\tau$ values and compare them with the results obtained using the maximum curvature. The results are presented in \cref{fig:13}. We observe that the same localization heads are consistently selected across different $\tau$ values, indicating that our analysis results are robust to the choice of $\tau$.

\noindent\textbf{Number of Heads $N$.} In Fig. 6(a) of the main paper, we select the 10 heads with the lowest $H(\mA^{\ell,h})$ values and repeat the process for 1,000 image-text pairs to calculate the selection frequency. We also investigate the effect of selecting different numbers of heads ($N$) on the selection frequency. We conduct experiments from $N = 1$ to $N = 14$ and compare the results with the selection frequency obtained using $N = 10$ (default setting). As shown in \cref{fig:14}, we can obtain the same top-3 localization heads consistently across different $N$ values, suggesting that the selection of localization heads is robust to the choice of $N$.
\section{More Experiments}
\subsection{Comparison with Baseline Models}
Most LVLMs, including the LLaVA~\cite{LLaVA} family, likely encode localization knowledge in their pretrained weights, possibly due to pretraining with bounding box coordinates or visual instruction prompts~\cite{LLaVA}. In~\cref{tab:6}, we compare baseline models and our proposed method, revealing the baseline models’ poor localization accuracy, likely due to their focus on describing objects rather than precise localization. Moreover, the localization head might provide only indirect support when text generation unfolds in its usual course. As a result, it becomes difficult for the model to directly output accurate object or region coordinates required for visual grounding, unless the information from this head is explicitly scrutinized. Thus, the localization head’s practical value can be realized as long as it is integrated with our proposed method.

\begin{table}[t]
    \centering
    \setlength{\tabcolsep}{3pt}
    \caption{Comparison performance to baseline models}
    \label{tab:6}
    \vspace{-5pt}
    \resizebox{\linewidth}{!}
    {
    \begin{tabular}{cccc}
        \Xhline{2\arrayrulewidth}
        \rowcolor{gray!20}
        { \textit{REC (RefCOCOg)}} 
        & {DeepSeekVL-1.3B} 
        & {LLaVA-1.5-7B} 
        & {LLaVA-1.5-13B} 
        \\
        \hline
        {Baseline}
        & {1.5} 
        & {2.92} 
        & {5.28} 
        \\
        {Ours}
        & {\textcolor{blue}{65.2}} 
        & {\textcolor{blue}{82.3}} 
        & {\textcolor{blue}{84.3}} 
        \\
        \Xhline{2\arrayrulewidth}
    \end{tabular}
    }
\vspace{-8pt}
\end{table}

\subsection{Comparision with F-LMM}

\begin{table}[t]
\setlength{\tabcolsep}{3pt}
\centering
\caption{Performance comparison between F-LMM~\cite{wu2024flmm} and our method on the RES task. We note that F-LMM models are trained on the training set of Referring Expression Segmentation datasets.}
\label{tab:7}
\vspace{-5pt}
\resizebox{\linewidth}{!}{
\begin{tabular}{lcccccccc}
\toprule
\multicolumn{1}{c}{\multirow{2}{*}{Method}} & \multicolumn{3}{c}{RefCOCO} & \multicolumn{3}{c}{RefCOCO+} & \multicolumn{2}{c}{RefCOCOg} \\
\cmidrule(lr){2-4}
\cmidrule(lr){5-7}
\cmidrule(lr){8-9}
 & val & testA & testB & val & testA & testB & val & test\\

\hline \hline

\rowcolor{gray!20}
\multicolumn{9}{l}{\textit{F-LMM (Fine-tuning on RES)}} \\
DeepSeek-VL-1.3B & 75.0 & 78.1 & 69.5 & 62.8 & 70.8 & 56.3 & 68.2 & 68.5  \\
Mini-Gemini-2B   & 75.0 & 78.6 & 69.3 & 63.7 & 71.4 & 53.3 & 67.3 & 67.4  \\
DeepSeek-VL-7B   & 76.1 & 78.8 & 72.0 & 66.4 & 73.2 & 57.6 & 70.1 & 70.4  \\
LLaVA-1.5-7B     & 75.2 & 79.1 & 71.9 & 63.7 & 71.8 & 54.7 & 67.1 & 68.1  \\

\hline \hline

\rowcolor{gray!20}
\multicolumn{9}{l}{\textit{Ours (Training-free)}} \\

DeepSeek-VL-1.3B & 56.3 & 57.0 & 52.7 & 51.2 & 55.5 & 49.2 & 52.3 & 55.8  \\
Mini-Gemini-2B   & 59.8 & 60.3 & 55.5 & 56.3 & 59.9 & 51.8 & 55.1 & 60.3  \\
DeepSeek-VL-7B   & 73.9 & 76.6 & 70.7 & 63.1 & 67.1 & 56.5 & 64.0 & 68.9  \\
LLaVA-1.5-7B     & 74.2 & 76.5 & 70.4 & 62.5 & 65.2 & 56.0 & 64.2 & 68.1  \\
  
\bottomrule

\end{tabular}

}
\vspace{-12pt}
\end{table}

We compare our method with F-LMM~\cite{wu2024flmm}, which also leverages the attention weights of frozen LVLMs for visual grounding. The differences between F-LMM and our method are as follows. First, F-LMM still requires fine-tuning its mask decoder modules on visual grounding datasets (i.e., referring expression segmentation datasets). Second, F-LMM uses all attention heads without considering the relative importance of each, leaving the decoder modules to interpret the entire set of attention weights. In contrast, our approach requires no fine-tuning and directly utilizes a few selected attention heads that are particularly useful for localizing objects in the image. Furthermore, our framework provides a transparent understanding of where the model focuses through localization heads, which is not available in F-LMM.

\cref{tab:7} presents the performance comparison between F-LMM and our method on the RES task. In smaller LVLMs (\eg, DeepSeek-VL-1.3B~\cite{lu2024deepseek} and Mini-Gemini-2B~\cite{li2024mini-gemini}), F-LMM outperforms our method. However, in relatively larger LVLMs (e.g., DeepSeek-VL-7B~\cite{lu2024deepseek} and LLaVA-1.5-7B~\cite{LLaVA1.5}), our method demonstrates performance comparable to F-LMM, with only a slight gap. This result suggests that the localization heads have competitive potential with the specialized mask decoder modules for visual grounding tasks, especially in relatively larger LVLMs.

\subsection{Gaussian Smoothing Ablation}
When assembling the attention map in the localization head (see Sec.~5 of the main paper), we apply Gaussian smoothing to the attention map to minimize potential random noise. In this section, we conduct an ablation study on the parameters of Gaussian smoothing to better understand the robustness of our framework across different values of standard deviation $\sigma$ and kernel size $\kappa$. For the experiments, LLaVA-1.5-13B~\cite{LLaVA1.5} was evaluated using the RefCOCO validation set (UNC split).

The results are presented in \cref{tab:8}. Regardless of the selected $\sigma$ and $\kappa$, Gaussian smoothing consistently enhances performance in almost all cases. The findings highlight that the framework is robust to varying choices of $\sigma$ and $\kappa$. Furthermore, even when using the basic attention map of localization heads without Gaussian smoothing ($\sigma = 0$ or $\kappa = 1$), the performance remains competitive, with only a 1.9\% drop compared to the best case. This demonstrates that Gaussian smoothing only serves as an auxiliary post-processing step for refining the attention map from localization heads.

\begin{table}[t]
\centering
\caption{Ablation study on Gaussian smoothing parameters ($\sigma$ and $\kappa$). The performance is evaluated using the RefCOCO validation set (UNC split) with the LLaVA-1.5-13B~\cite{LLaVA1.5}.}\label{tab:8}
\vspace{-5pt}
\resizebox{0.8\linewidth}{!}{
\begin{tabular}{ccccccc}
\toprule
\multirow{2}{*}{Task} & \multicolumn{6}{c}{$\sigma$ (standard deviation)} \\
\cline{2-7}
& 0.0 & 0.4 & 0.8 & 1.0 & 1.4 & 1.8 \\
\hline
REC & 85.5 & 86.8 & 87.2 & \textcolor{blue}{87.2} & 86.8 & 84.3 \\
RES & 74.3 & 75.2 & 76.1 & \textcolor{blue}{76.1} & 75.2 & 72.7 \\
\midrule
\multirow{2}{*}{Task} & \multicolumn{6}{c}{$\kappa$ (kernel size)} \\
\cline{2-7}
& 1 & 3 & 5 & 7 & 9 & 11 \\
\hline
REC & 85.5 & 86.5 & 86.5 & \textcolor{blue}{87.2} & 87.2 & 87.2 \\
RES & 74.3 & 75.2 & 75.2 & \textcolor{blue}{76.1} & 76.1 & 76.1 \\
\bottomrule
\end{tabular}}
\vspace{-6pt}
\end{table}

\begin{table}[t]
    \centering
    \caption{Performance comparison with F-LMM~\cite{wu2024flmm} on the PNG~\cite{gonzalez2021panoptic} benchmark.}
    \label{tab:9}
    \vspace{-5pt}
    \resizebox{0.60\linewidth}{!}{
    \begin{tabular}{cccc}
    \Xhline{2\arrayrulewidth}
    \rowcolor{gray!20}
    {\textit{PNG (all)}} & 
    {Ours} & 
    {F-LMM} \\
    \Xhline{1\arrayrulewidth}
    {DeepSeekVL-7B} & 
    {\textcolor{blue}{66.7}} & 
    {65.7} \\
    \Xhline{2\arrayrulewidth}
    \end{tabular}
    }
\vspace{-12pt}
\end{table}

\subsection{Multi-Object Grounding Tasks}
Beyond single-object tasks, our pipeline also suggests promise for multi-object grounding. 
We utilize spaCy~\cite{spacy} to extract noun tokens for generating attention maps (see \cref{fig:10}), obtaining comparable results on the PNG benchmark~\cite{gonzalez2021panoptic}, with improvements observed relative to F-LMM (see \cref{tab:9}).
Similarly, we believe this approach holds promise for extension to other various tasks~\cite{he2023grec,liu2023gres,rasheed2024glamm}.

\section{More Qualitative Results}
We present more qualitative results of our framework, including the performance of 10 LVLMs~\cite{LLaVA,LLaVA1.5,lu2024deepseek,li2024mini-gemini,chen2024internvl,chen2023sharegpt4v,young2024yi}, with parameter numbers ranging from 1.3B to 13B, on visual grounding tasks. \cref{fig:15}, \cref{fig:16}, and \cref{fig:17} present the qualitative results of our method on the Referring Expression Comprehension (REC), Referring Expression Segmentation (RES), and Reasoning Segmentation tasks, respectively. The results demonstrate that only a few selected localization heads are sufficient to accurately localize objects in the image based on the text query. Our method effectively localizes objects in various scenarios.
\section{Applications}
\subsection{Real World Application}
\cref{fig:18} illustrates that the localization heads effectively capture the region or object of interest in images from the real world, based on the provided expressions. This result demonstrates the robustness of the localization heads across various types of data.

\subsection{Image Editing}
\cref{fig:19} presents the results of image inpainting performed by integrating Stable Diffusion XL (SDXL)~\cite{podellsdxl}. The frozen LVLM generates a segmentation mask corresponding to the expression, and this mask, along with an additional text prompt, is used as input to the diffusion model to generate the desired image. These results demonstrate that the segmentation mask corresponding to the referred text, output by a small number of localization heads from the frozen LVLM, can serve as guidance for diffusion models. This compatibility enables its application in image editing tasks.
\section{Limitations}
We propose a simple yet effective framework for training-free visual grounding, which leverages the localization heads of LVLMs. Our framework successfully localizes objects in images based on text queries without requiring any fine-tuning and achieves superior performance compared to existing training-free methods. However, our method still has some limitations that could be addressed in future work.

First, our work, as illustrated in ~\cref{fig:10}, reveals the potential for multi-object grounding; however, the establishment of a formalized pipeline or the development of a more streamlined implementation remains limited. The task of rendering the identified localization head more practical, user-friendly, and adaptable across a diverse range of applications continues to pose a significant challenge. This presents a compelling avenue for future research.

Second, our method is less suitable for LVLMs or methods that do not preserve spatial information in images (\eg, pooling)~\cite{BLIP2, qwenvl, laurencon2023obelics, internvl2, jaegle2021perceiver}. These methods make it challenging to explicitly obtain image attention maps. To collect the attention map, a reverse computation is required to determine the order in which image tokens were input during processing. We leave the application of our framework to these methods for future exploration.
\begin{figure}[t]
    \centering
    \includegraphics[width=\linewidth]{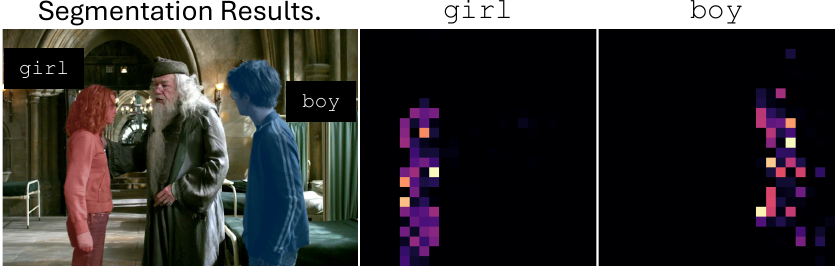}
    \caption{Multi-object segmentation results from the localization heads of DeepSeekVL-7B, along with the corresponding raw attention maps.}
    \label{fig:10}
\end{figure}
\clearpage\newpage
\begin{figure*}[t]
    \centering
    \includegraphics[width=0.7\linewidth]{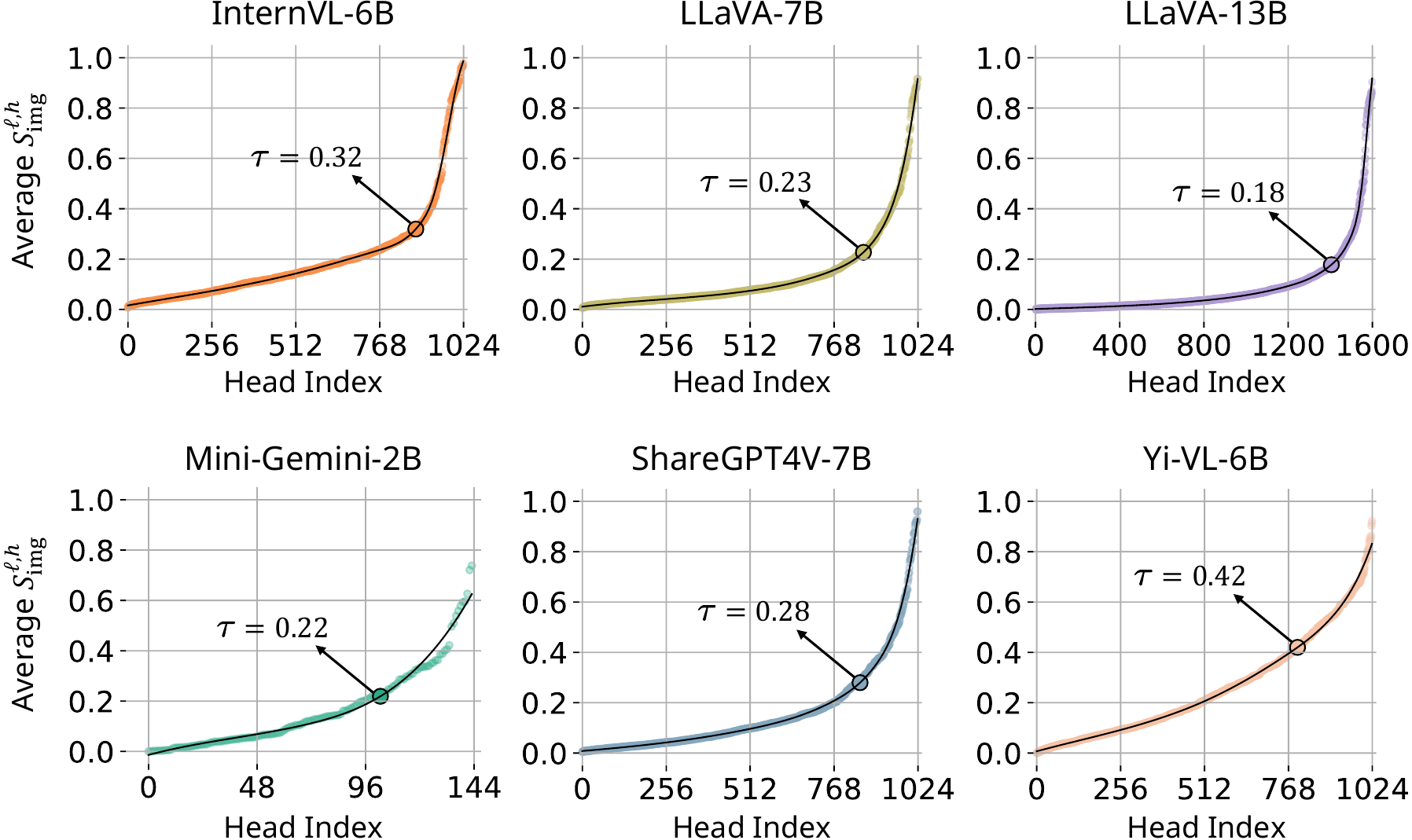}
    \caption{Average $S_{\text{img}}^{\ell, h}$ values for each attention head in more LVLMs. $\tau$ is set at the point where the maximum curvature is observed.}
    \label{fig:11}
\end{figure*}

\begin{figure*}[t]
    \centering
    \includegraphics[width=0.8\linewidth]{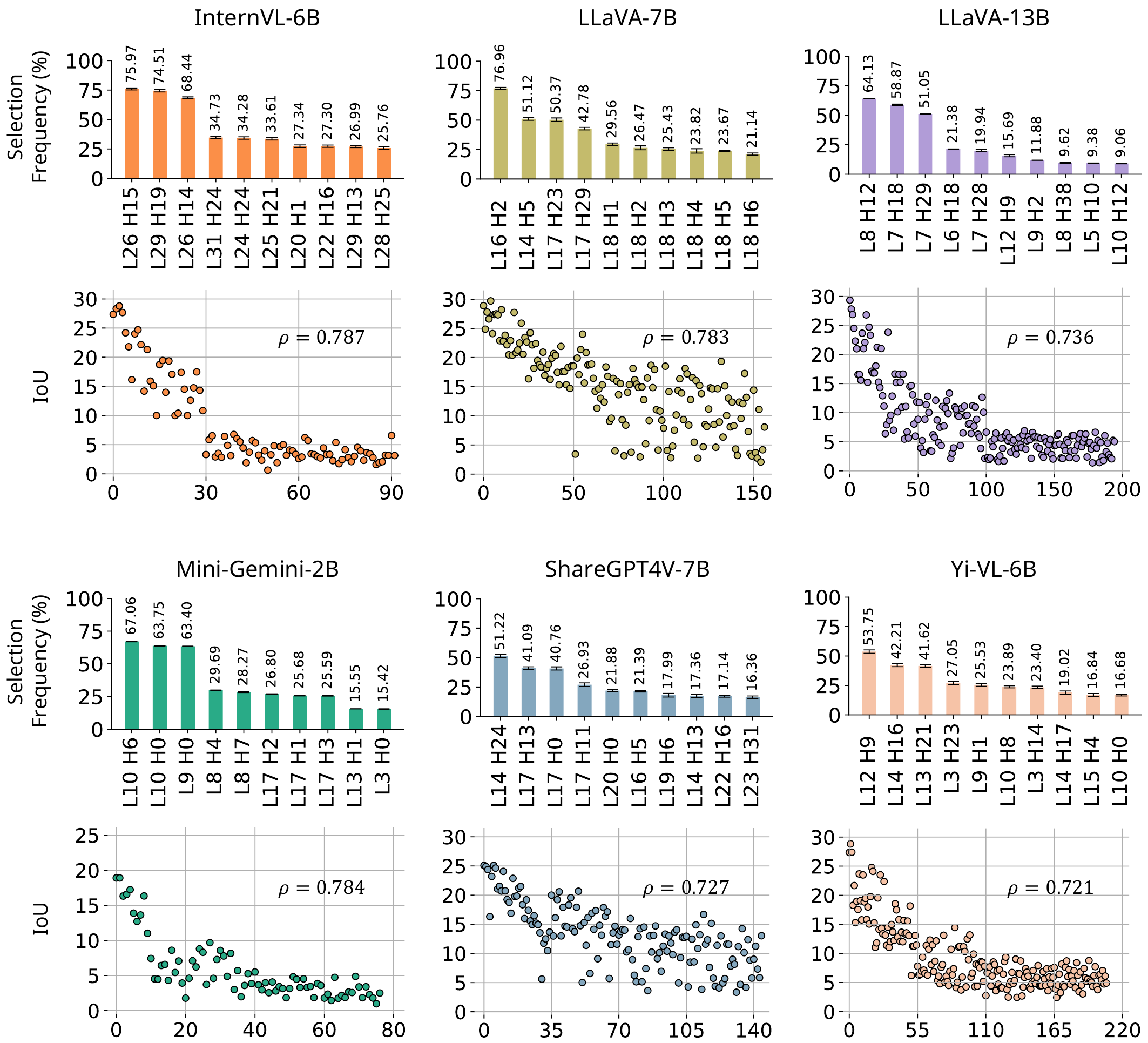}
    \caption{Selection frequency of individual heads and scatter plot of selection frequency rank versus each head's average IoU in more LVLMs. The Spearman correlation coefficient ($\rho$) between the selection frequency rank and the average IoU is displayed in the top-right corner of each plot. The observed Spearman correlation are statistically significant ($p < 0.001$) for all LVLMs.}
    \label{fig:12}
\end{figure*}
\begin{figure*}[h]
    \centering
    \includegraphics[width=0.54\linewidth]{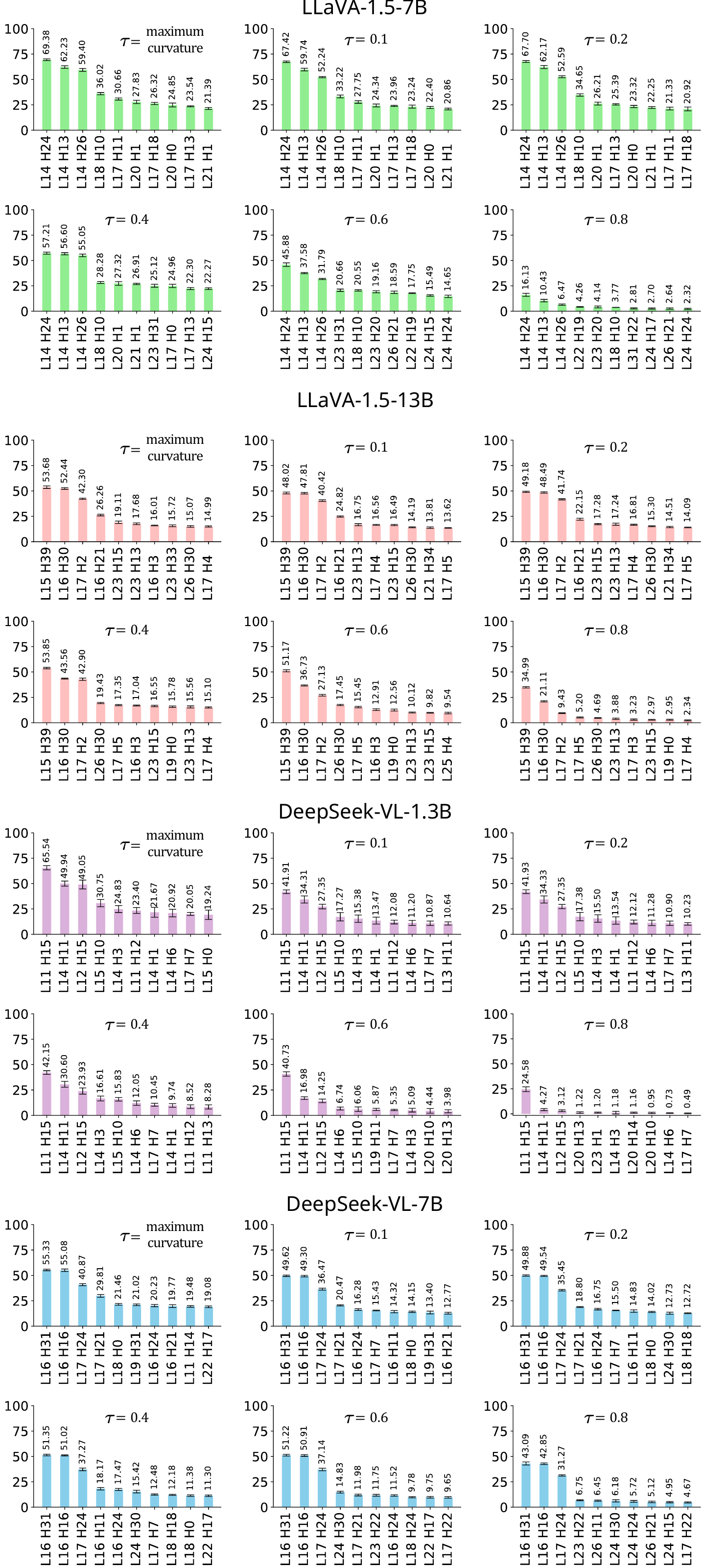}
    \caption{Selection frequency of individual heads across different $\tau$ values. $\tau$ represents the threshold for the sum of each head's attention map. Our analysis focuses on heads with attention map sums greater than $\tau$, which are selected as targets for selection frequency evaluation. In the main paper, we select the threshold where the maximum curvature is observed. The top-3 localization heads remain consistent across different $\tau$ values, demonstrating the robustness of our analysis to variations in $\tau$.}
    \label{fig:13}
    \vspace{-10pt}
\end{figure*}
\begin{figure*}[h]
    \centering
    \includegraphics[width=0.77\linewidth]{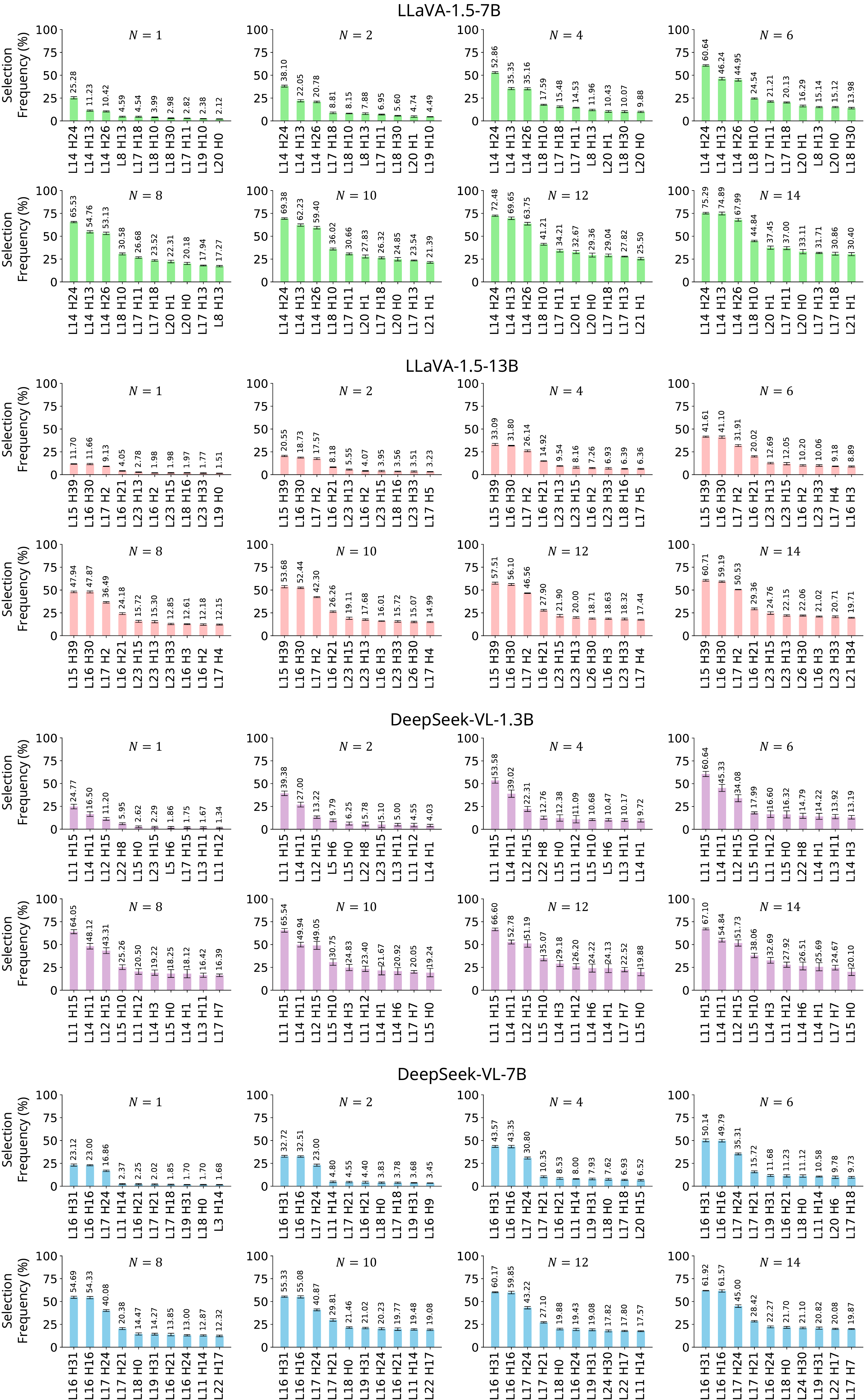}
    \caption{Selection frequency of individual heads across different $N$ values. $N$ refers to the number of selected heads based on the lowest $H(\mA^{\ell,h})$ values. Default setting is $N = 10$. The top-3 localization heads are consistent across different $N$ values, indicating the robustness of localization head selection to the choice of $N$.}
    \label{fig:14}
\end{figure*}

\begin{figure*}[h]
    \centering
    \includegraphics[width=0.69\linewidth]{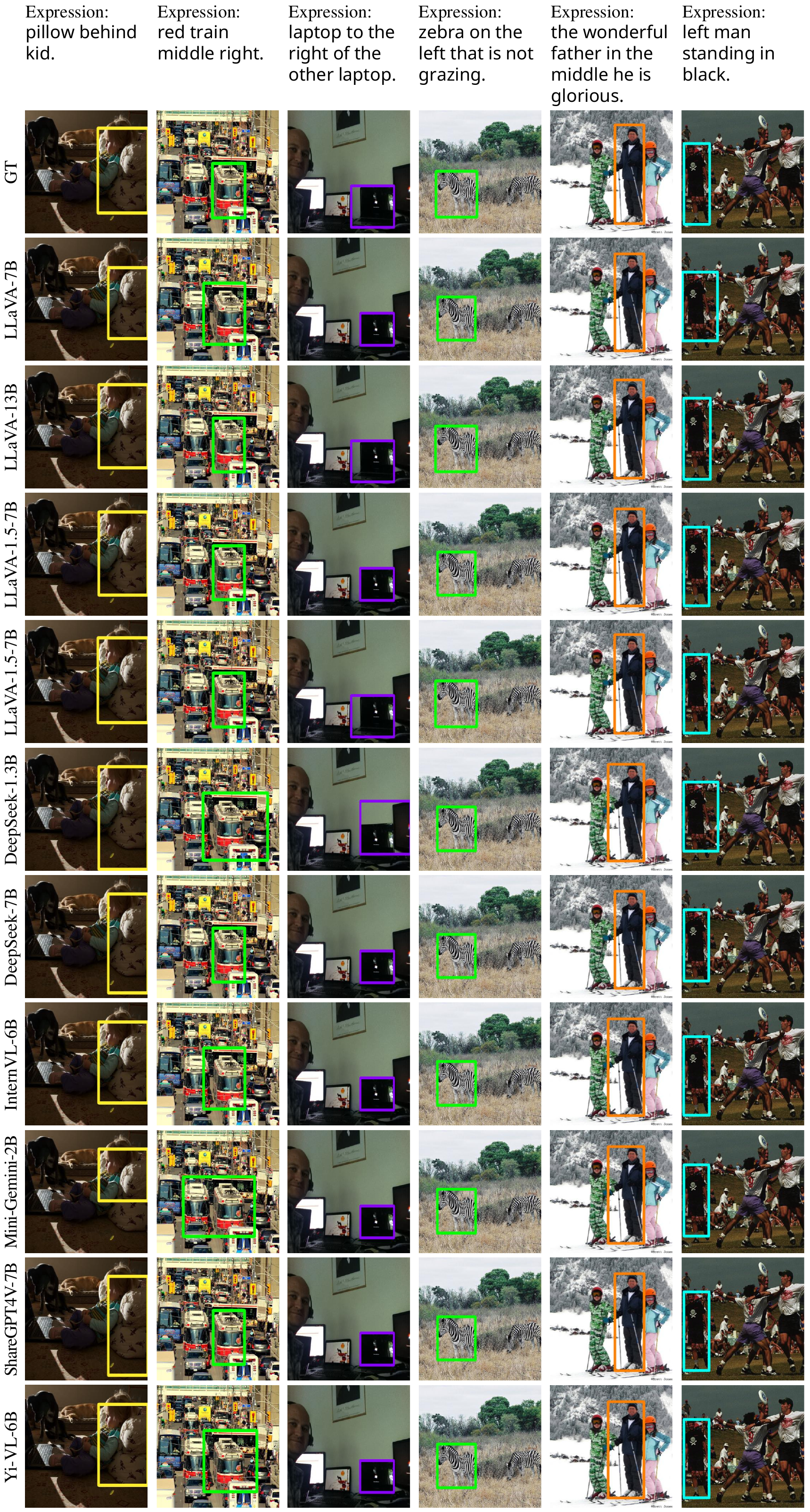}
    \caption{Qualitative results of Referring Expression Comprehension.}
    \label{fig:15}
\end{figure*}
\begin{figure*}[h]
    \centering
    \includegraphics[width=0.70\linewidth]{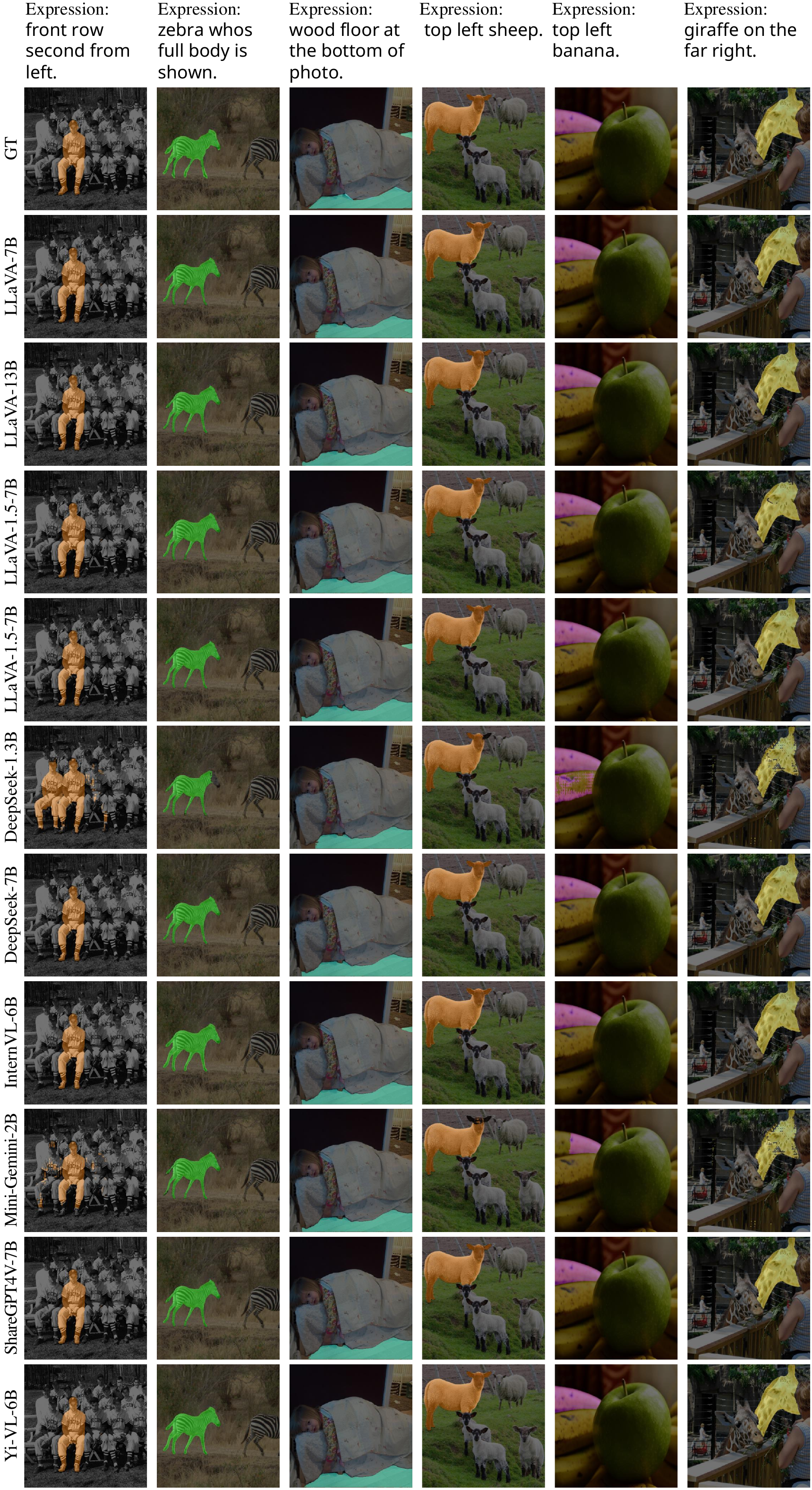}
    \caption{Qualitative results of Referring Expression Segmentation.}
    \label{fig:16}
\end{figure*}
\begin{figure*}[h]
    \centering
    \includegraphics[width=\linewidth]{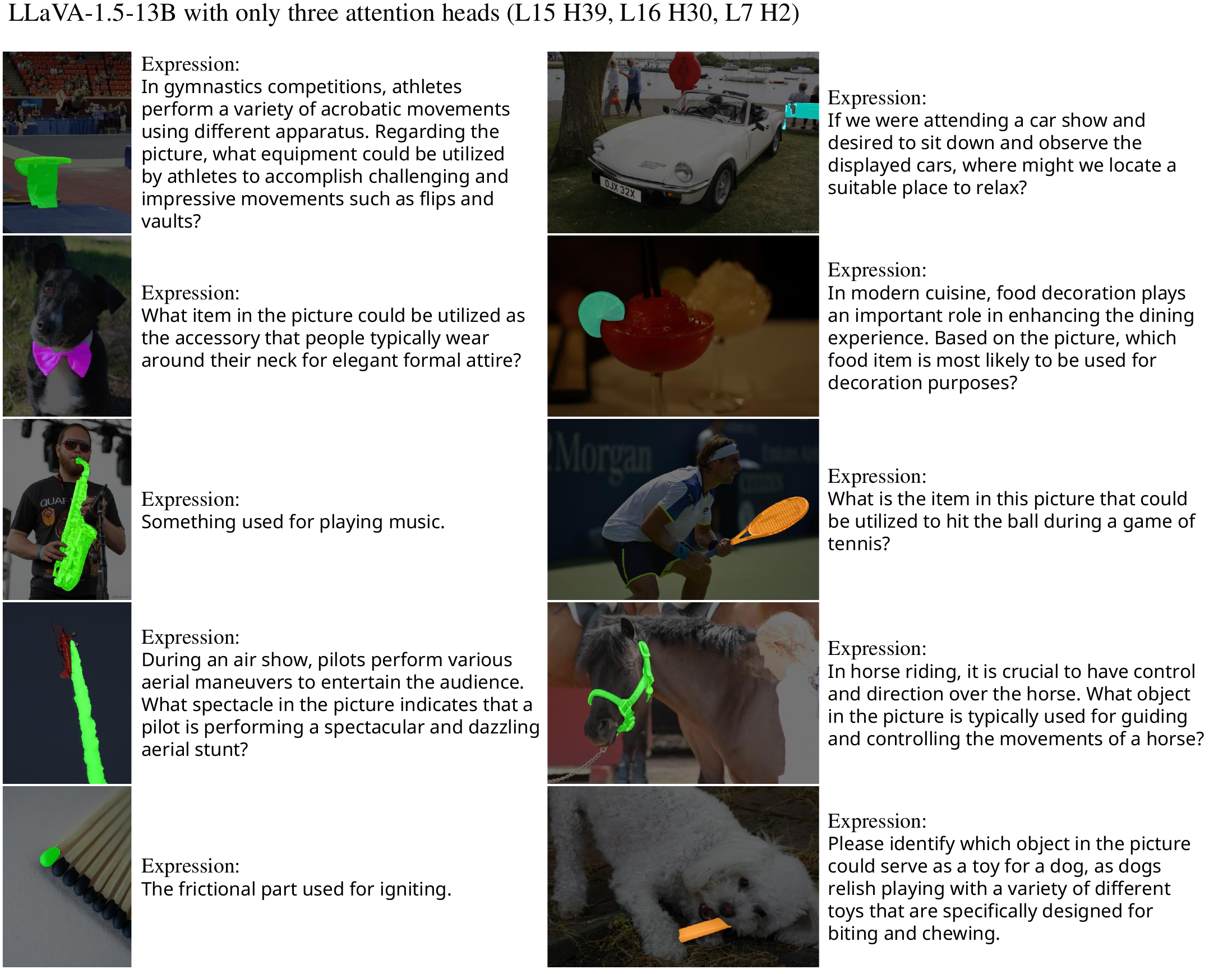}
    \caption{Qualitative results of Reasoning Segmentation.}
    \label{fig:17}
\end{figure*}
\begin{figure*}[h]
    \centering
    \includegraphics[width=0.84\linewidth]{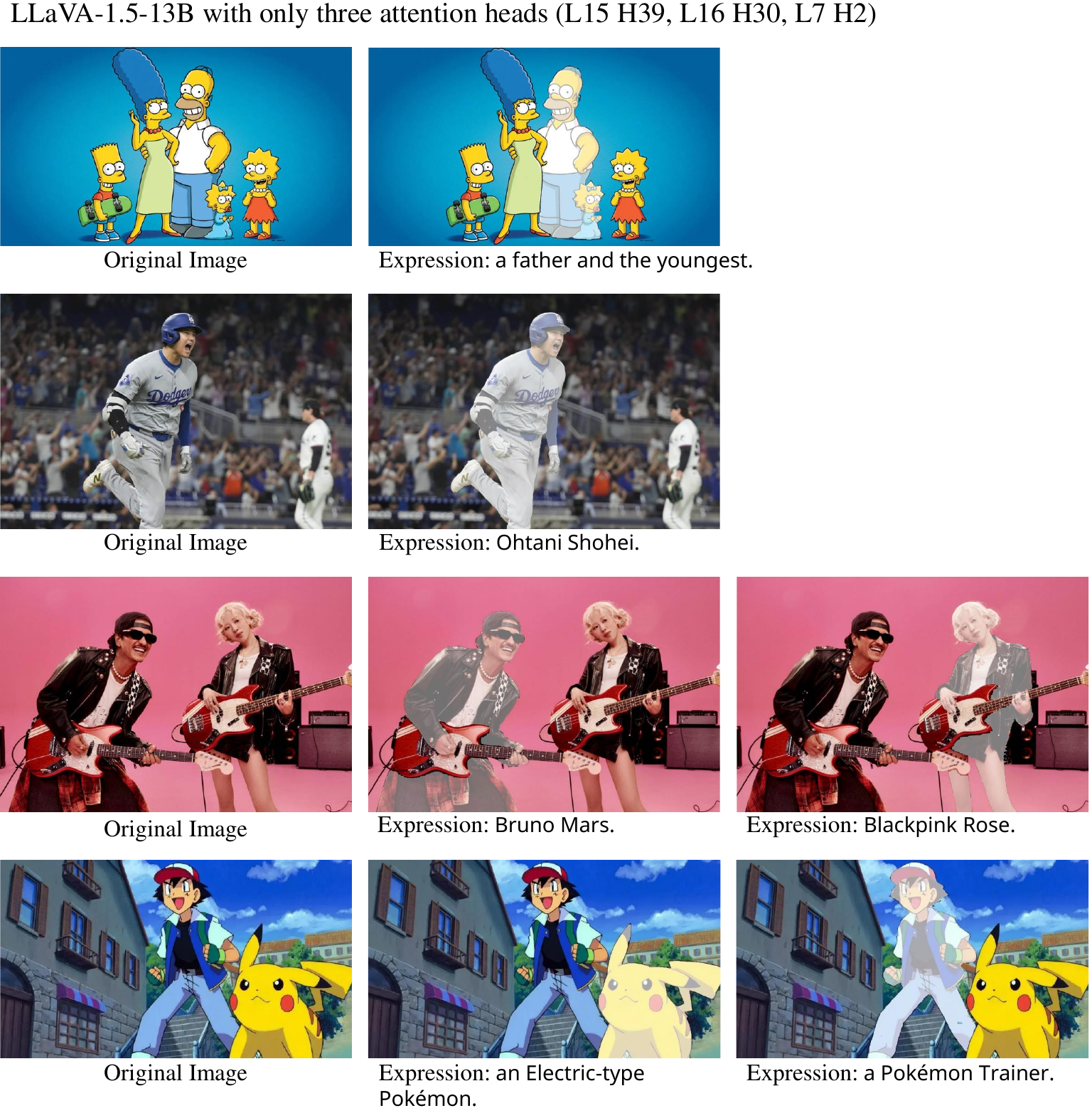}
    \caption{Qualitative results of real-world image segmentation. LLaVA-1.5-13B~\cite{LLaVA1.5} uses only three attention heads (L15 H39, L16 H30, L7 H2) as localization heads to produce a precise segmentation masks related to the text expressions. The whitened regions in the images represent the segmentation mask output by the model.}
    \label{fig:18}
\end{figure*}
\begin{figure*}[h]
    \centering
    \includegraphics[width=0.88\linewidth]{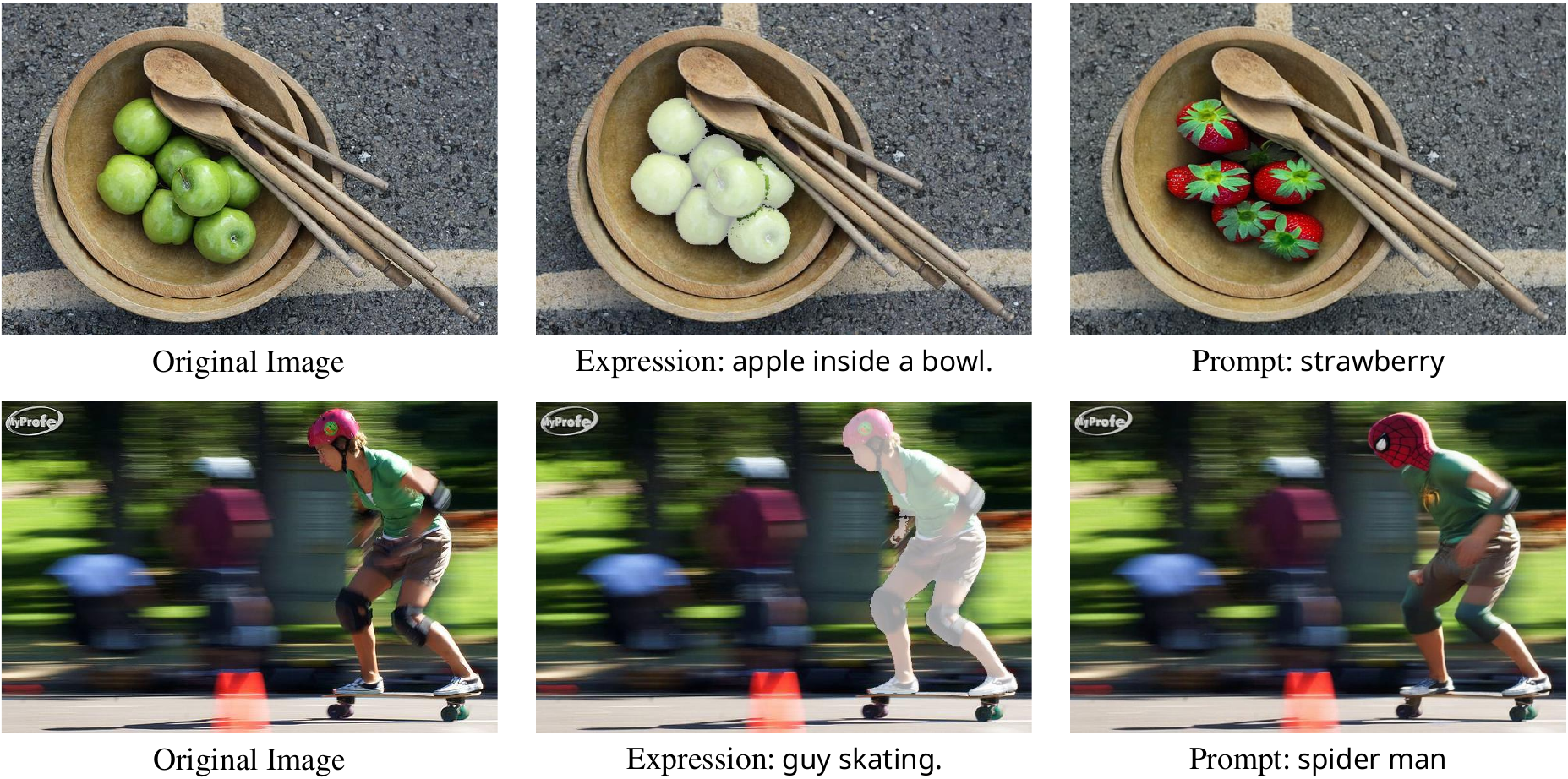}
    \caption{Qualitative results of generating the desired image through integration with a diffusion model~\cite{podellsdxl}. Given an original image, our method generates a mask from the LVLM based on the text describing the desired modifications. This mask is then used as guidance for a diffusion model to perform image editing. Using the segmentation mask obtained through the localization head of the frozen LVLM~\cite{LLaVA1.5}, it is possible to generate semantic objects that align with the prompt at the specified mask locations.}
    \label{fig:19}
\end{figure*}

\end{document}